\documentclass[journal]{IEEEtran}
\usepackage[numbers,sort&compress]{natbib}

\usepackage{amsmath}
\usepackage{amssymb}

\usepackage{amsthm}
\newtheorem{proposition}{Proposition}
\usepackage{algorithmic}
\usepackage{algorithm}
\usepackage[caption=false,font=footnotesize]{subfig}
\usepackage[percent]{overpic}
\usepackage{mathtools}
\usepackage{bbm}
\usepackage{booktabs}
\usepackage{placeins}
\usepackage{array}
\usepackage{graphicx}
\usepackage[dvipsnames]{xcolor}
\usepackage{hyperref}
\usepackage{enumitem}
\usepackage{wrapfig}
\hypersetup{
    colorlinks=true,
    linkcolor=NavyBlue,
    filecolor=magenta,      
    urlcolor=NavyBlue,
    citecolor=NavyBlue,
    }

\definecolor{robotgreen}{HTML}{7FC97F}
\definecolor{handbg}{HTML}{FDF3E1}\definecolor{handfill}{HTML}{F9D9A6}\definecolor{handdark}{HTML}{D98B1E}
\definecolor{robotbg}{HTML}{EAF5E6}\definecolor{robotfill}{HTML}{C9E8C0}\definecolor{robotdark}{HTML}{4E9A3F}
\definecolor{gatefill}{HTML}{DCE8F7}\definecolor{gatedark}{HTML}{3E6FB0}
\definecolor{gapred}{HTML}{bd0808}

\ExplSyntaxOn
\newcommand{\plus}[1]{
  \fp_compare:nTF { #1 >= 10 }
    {({\color{blue} \textbf{+#1}})} %
    {({\color{blue} +#1})}          %
}
\ExplSyntaxOff

\ExplSyntaxOn
\newcommand{\minus}[1]{
  \fp_compare:nTF { #1 >= 10 }
    {({\color{red} \textbf{-#1}})} %
    {({\color{red} -#1})}          %
}
\ExplSyntaxOff

\newcommand*\circled[1]{\raisebox{.5pt}{\textcircled{\raisebox{-.9pt}{\scriptsize #1}}}}

\title{DUO Policy: Bridging Human Video Coverage and Robot Precision for Fine-Grained Manipulation}

\newif\ifanon
\anonfalse

\ifanon
\author{Anonymous Authors}
\else
\author{Mingxi Jia$^{1}$, Patton Jiyang Yin$^{1}$, Haojie Huang$^{2}$, Ivy Xiao He$^{1}$, Xuanru Liu$^{1}$, Gefei Liu$^{1}$, \\ Dian Wang$^{3}$, and Stefanie Tellex$^{1}$\\[2pt]
{\normalsize $^{1}$Brown University\quad $^{2}$Northeastern University\quad $^{3}$Stanford University}}
\IEEEaftertitletext{\vspace{-1.2\baselineskip}}
\fi

\begin{document}
\bstctlcite{IEEEexample:BSTcontrol}
\maketitle

\begin{abstract}
Human videos promise a scalable way to reduce expensive robot data collection, but the embodiment gap limits the resulting policies to coarse behaviors, and they fail at precise tasks such as insertion. Existing methods co-train or finetune a single policy on both data sources, treating each source as equally useful in every state: the policy either absorbs the embodiment error at precision-critical steps or spends scarce robot data relearning free-space motion. We observe that the value of each source depends on the state: human data is useful where precision requirements are low, while robot data is disproportionately valuable near precision-critical states. We therefore propose \textbf{DUO Policy} (Dual-Actor Policy), which decouples a manipulation skill into a hand policy, pretrained on human videos for global free-space motion, and a robot policy, trained on sparse in-hand data for local precise corrections, with a gate learned from human interventions switching between them. This design is robust to moderate embodiment error in human data and generalizes to scene layouts beyond the coverage of the robot data. On tasks such as K-cup coffee making, desk cleanup, and cluttered candy sorting, DUO Policy outperforms the baseline with the highest average success by 12.9 percentage points over four simulated tasks and by 51.7 points over three real-world tasks, and matches robot-only training with 2.75$\times$ less data-collection time.\ifanon\else{} Project page: \url{https://mingxi-jia.github.io/duo_policy_site/}.\fi

\end{abstract}


\section{Introduction}

Cross-embodiment learning promises a scalable solution to use data from other embodiments to break the data bottleneck in imitation learning. Human videos in particular are cheap to collect, can cover the workspace densely, and capture a diverse set of behaviors~\citep{phantom,masquerade,kareer2025egomimic}. However, policies learned from these data sources often fail on fine-grained manipulation tasks such as tight-fit insertion, even when co-trained with robot data, because the observation and retargeted human motions differ systematically from what the target robot should do. Human videos can still benefit fine-grained manipulation, provided each data source is used where it is reliable.

\begin{figure}[!t]
    \centering
    \includegraphics[width=\linewidth]{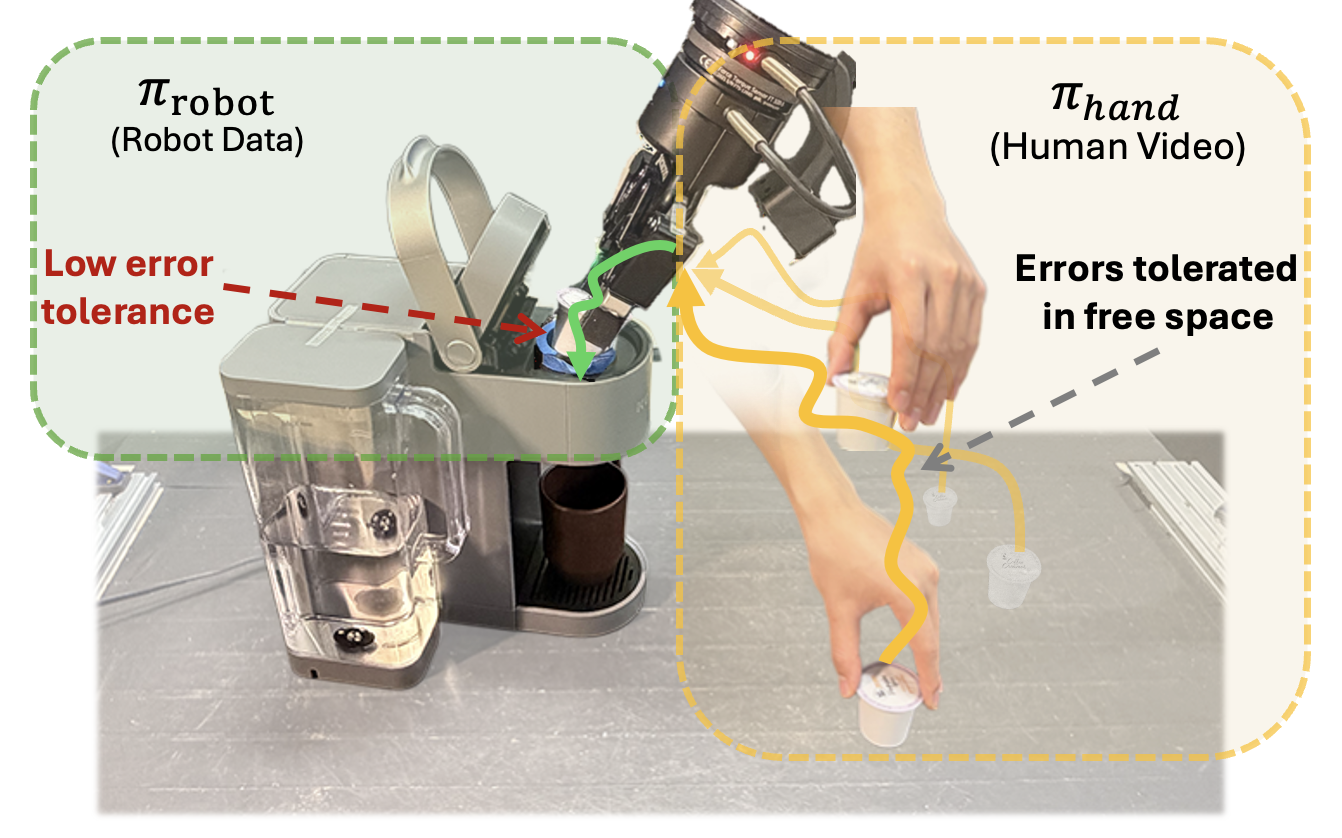}
    \caption{\textbf{Where Is Each Data Source Useful?} \textcolor{orange}{Human Hand Data}
is imprecise but sufficient in free space. Near insertion, only
\textcolor{robotgreen}{Robot Data} is precise enough. DUO Policy uses each
where it is valuable.}
\label{fig:intro_example}
\end{figure}

\begin{figure*}[t]
    \centering
    \subfloat[Dataset Coverage]{\includegraphics[width=0.33\linewidth]{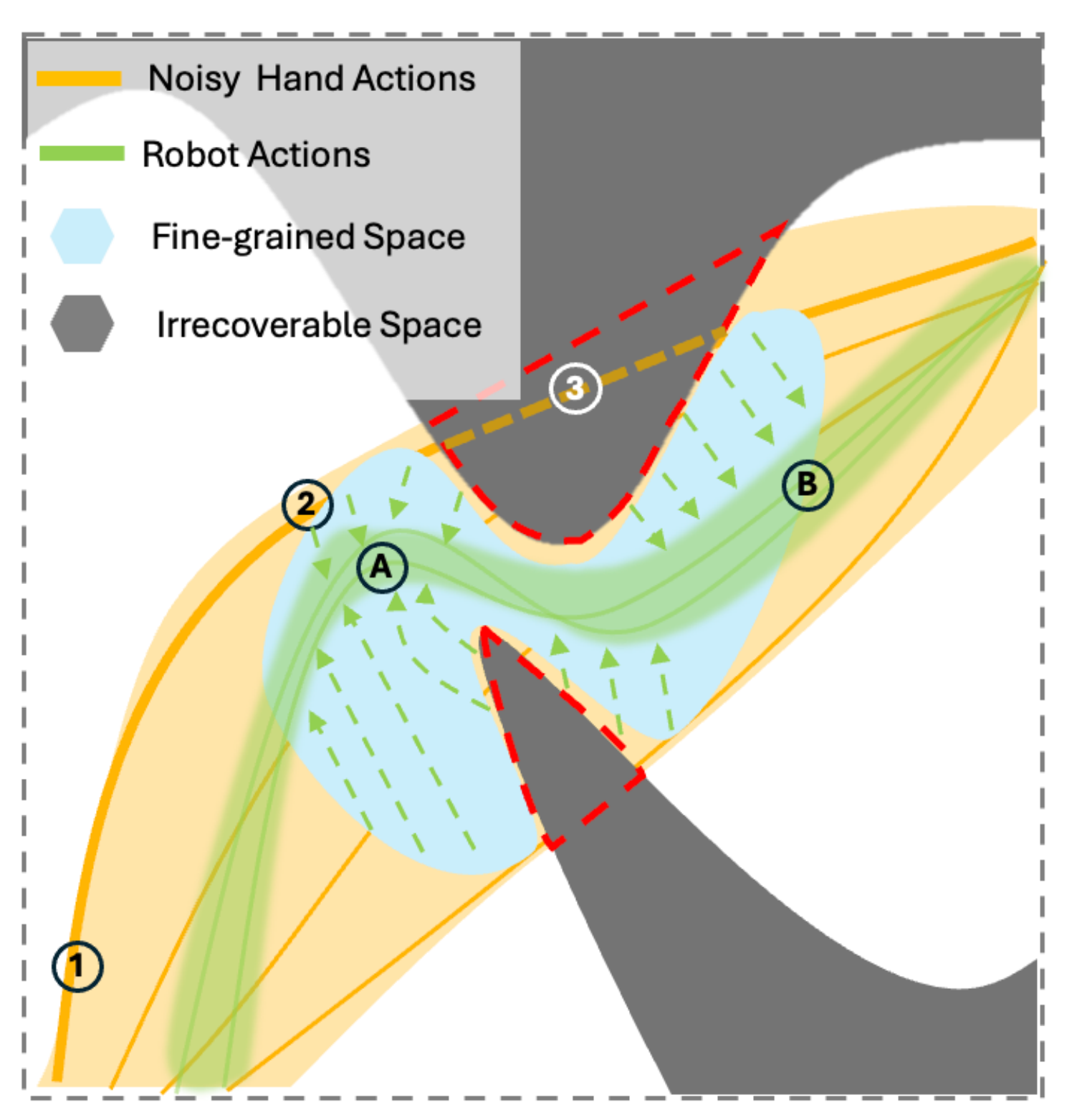}\label{fig:why_a}}\hfill
    \subfloat[Policy Behaviors]{\includegraphics[width=0.67\linewidth]{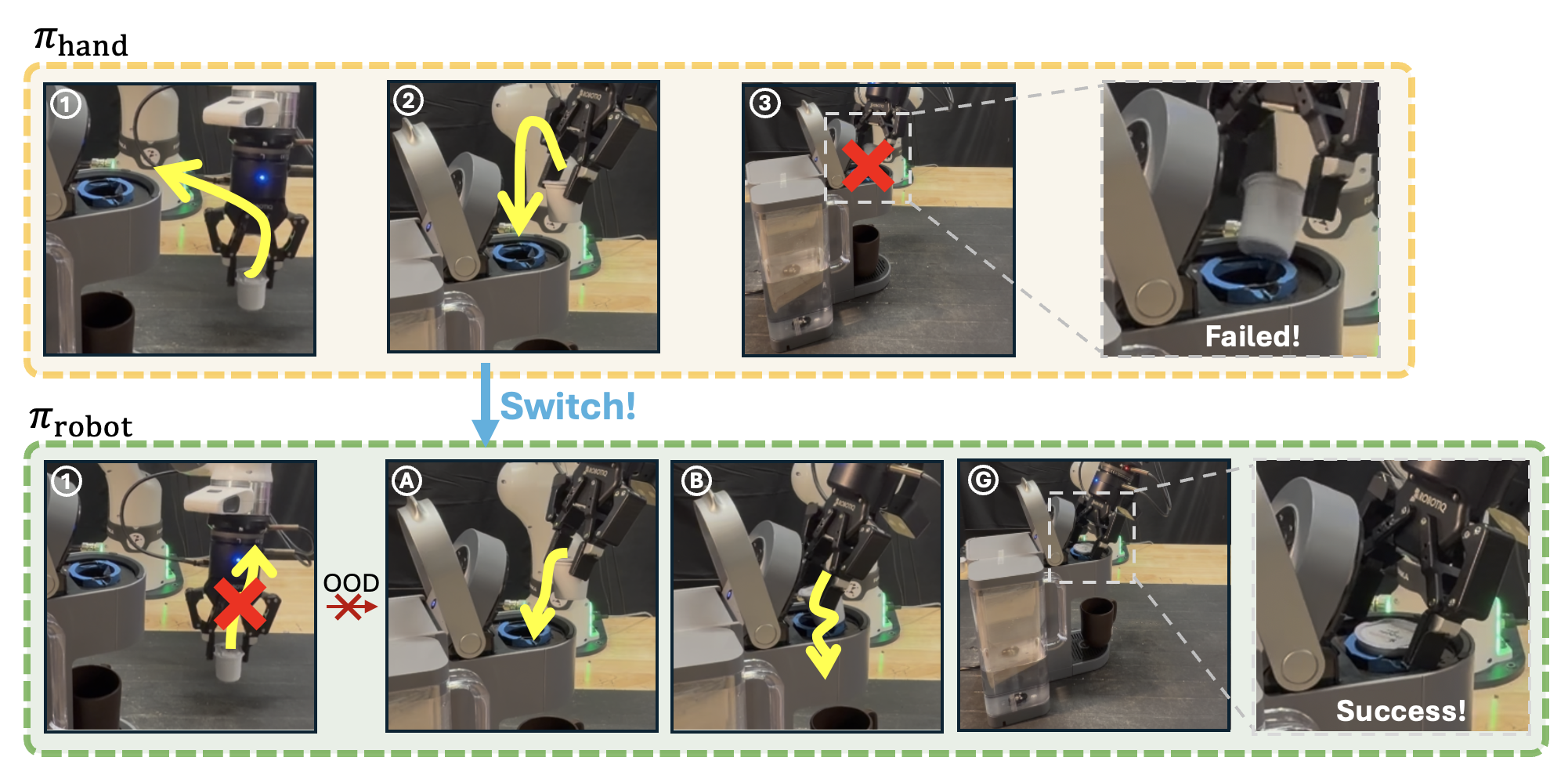}\label{fig:why_b}}
    
    \caption{\textbf{Each Data Source Is Valuable in Different States.} \textbf{(a)} \textcolor{orange}{Human hand data} covers the workspace, but its embodiment error leads into \textcolor{gray}{irrecoverable states} in the narrow precision-critical region (blue). \textcolor{robotgreen}{Robot data} is precise but sparse. \textbf{(b)} $\pi_{\text{hand}}$ collides during insertion~\circled{3}, and $\pi_{\text{robot}}$ fails from an out-of-distribution start~\circled{1}. DUO Policy follows $\pi_{\text{hand}}$ through free space~\circled{1}--\circled{2}, switches to $\pi_{\text{robot}}$ near the holder~\circled{A}--\circled{B}, and completes the insertion~\circled{G}.}
  \label{fig:why}
\end{figure*}

In this paper, we argue that \textbf{the value of cross-embodiment data depends on the state, and is governed by how much precision that state demands.} Let
$\varepsilon(s)$ denote the largest action error from which the task can still succeed at state $s$. Cross-embodiment data introduces a systematic action error $\eta$. Where $|\eta| < \varepsilon(s)$, as in reaching, transporting, and approaching, the error is absorbed and cross-embodiment data contributes to better coverage where robot data cannot cheaply provide. However, when $\varepsilon(s)$ is below $|\eta|$, as in aligning, inserting, and closing a contact, cross-embodiment data becomes actively misleading, and target-robot data can be worth more than far larger amounts of human video. In short, \emph{cross-embodiment data is useful where precision requirements are low. Target-embodiment or robot data is disproportionately valuable near precision-critical states.}

This view explains why existing recipes succeed in general manipulation tasks but struggle with fine-grained manipulation. Phantom~\citep{phantom} fills the visual gap between human and robot by inpainting, but the gap remains in the action space. Co-training methods~\citep{masquerade,kareer2025egomimic} treat each data source as uniformly useful across states, which confuses the two sources in precision-critical states. Target-supervised finetuning~\citep{yang2025egovla,wang2024hpt} adapts the pretrained model with full robot demonstrations, so scarce robot data must also cover free-space behavior the cross-embodiment prior already provides. When the model~\citep{yang2025egovla} is fully finetuned, it also risks forgetting that prior.

We turn this principle into \textbf{DUO Policy} (Dual-Actor Policy). A hand policy pretrained on human video handles the low-precision portion of the task. A robot policy trained on sparse, in-hand robot data handles the precision-critical portion. A learned policy gate switches between them. The gate does not need hand-designed phase labels. We collect robot data with human-in-the-loop interventions, and the moments where an operator takes over mark the states where the human prior's error exceeds the tolerance. Intervention labels therefore serve as a free proxy for precision-criticality, and the operator's effort is spent where it matters. Our contributions can be summarized as follows:
  \begin{itemize}[leftmargin=10pt, labelindent=0pt]
    \item \textbf{State-dependent data value.} We show that the value of cross-embodiment data depends on the state: it helps where precision requirements are low and misleads near precision-critical states, where sparse robot data is most valuable.
    \item \textbf{Dual-actor architecture.} We introduce DUO Policy, a dual-actor architecture that learns global motion from human videos and local corrections from sparse robot data, with a gate trained on human interventions that switches between the two actors.
    \item \textbf{Equivariance by construction.} We show that separating the global and local observation and action spaces makes DUO Policy equivariant to planar rigid transformations of the scene in precision-critical states, so robot corrections learned at one placement transfer to others.
    \item \textbf{Empirical gains.} We validate our method through extensive experiments, outperforming the baseline with the highest average success by 12.9 percentage points over four simulated tasks and 51.7 points over three real-world tasks, with $2.75\times$ less data-collection time on candy sorting.
  \end{itemize}

\section{Related Work}

\textbf{Visuomotor Policy Learning:} Behavior cloning from robot demonstrations~\cite{robomimic, diffpo, zhao2023learning} learns precise skills but needs many demonstrations; data efficiency improves with 3D representations~\citep{dp3}, symmetry~\citep{so2rl, huang2024leveraging, equidiff, hu20263d, jia2025learning, wang2026equivariant, huang2026pix2act, huang2026action}, data augmentation~\citep{jia2023seil, papagiannis2025miles}, and hybrid action modes~\citep{belkhale2023hydra}. Generalist policies pretrained on large multi-robot datasets~\citep{octo2024, kim2024openvla, black2025pi0, wang2024hpt} still need target-robot demonstrations covering the whole task. DUO instead learns the free-space motion from human videos, so robot data needs to cover only the precision-critical states.

\textbf{Learning from Human Videos}: Human videos~\citep{grauman2022ego4d, hoque2025egodex, fu2025gigahands, chen2025foundhand, min2024genheld, li2026deform360, min2026motionsplicer, li2026novaflow} are a valuable alternative data source for learning visuomotor policies~\citep{wang2023mimicplay, kareer2025egomimic, zheng2026egoscale, yang2025egovla}. Phantom~\citep{phantom} uses human videos by overlaying rendered robot arms on RGB frames. Motion Tracks~\cite{ren2025motion} uses keypoints as an intermediate representation to align human and robot actions. Masquerade~\citep{masquerade} edits human videos to look like robot data, pretrains a visual encoder to predict future hand keypoints, and keeps this prediction as an auxiliary loss while training robot actions. MimicPlay~\citep{wang2023mimicplay} learns latent plans from human play data to guide a low-level robot policy. These methods use human data throughout the task, so its embodiment error also reaches the precision-critical steps. DUO uses human data only where this error is tolerable: a hand policy trained on human videos handles free-space motion, and a robot policy trained on in-hand robot data takes over at precision-critical states.

\textbf{Human-in-the-Loop Policy Learning}: Human corrections mitigate covariate shift in imitation learning~\citep{ross2011reduction, kelly2019hgdagger, korkmaz2025mile}. Sirius~\citep{liu2025robot} upweights corrective samples in behavior cloning, and CR-DAgger~\citep{xu2026compliant} learns a residual policy on top of a base policy. These methods use corrections to refine a policy within one embodiment. DUO uses interventions for cross-embodiment transfer: they mark where the policy trained on human data fails, train the gate that hands control to the robot policy at these states, and provide the corrections the robot policy learns from.

\section{Problem Setting}
\label{sec:problem}
We consider learning a fine-grained manipulation policy from two heterogeneous sources: a large, low-cost human-video dataset $\mathcal{D}_h = \{\tau_i^h\}_{i=1}^{N_h}$ and a small, interactive robot dataset $\mathcal{D}_r = \{\tau_j^r\}_{j=1}^{N_r}$. Unlike robot data, human videos lack in-hand views. After retargeting human hand motion into the robot's end-effector action space, we view $\mathcal{D}_h$ as demonstrations from a biased expert whose retargeted actions satisfy $\hat{a}_t = \pi^*(s_t) \oplus \eta(s_t)$, where $\oplus$ applies the systematic cross-domain pose offset $\eta$. The same retargeting error also enters the observations, since the end-effector pose in $o_t$ and the gripper rendered into the point cloud are placed at the retargeted pose. Each robot trajectory $\tau_j^r = \{(o_t, a_t, i_t)\}$ is collected interactively, where the intervention label $i_t \in \{0,1\}$ marks the fine-grained timesteps at which a human takes over. $a_t = a_t^*$ is the operator's precise action when $i_t=1$, and $a_t$ is the action of the deployed hand policy when $i_t=0$. Our objective is to learn a policy that maximizes task success while minimizing robot-data collection time.
Let $\varepsilon(s)$ be the precision tolerance at state $s$, i.e., the largest deviation of the commanded end-effector pose from the expert's from which the task can still succeed. We probe the unobserved $\varepsilon(s)$ with a sensitivity test that perturbs expert actions at sampled states and measures the drop in success (Appendix~\ref{appendix:extra}). We call $\mathcal{S}_{\text{crit}} = \{s : \varepsilon(s) < |\eta(s)|\}$ the \emph{precision-critical} states, and the remaining states free space.

\section{Method}

DUO Policy builds on the key observation illustrated in Fig.~\ref{fig:why}: a fine-grained task often combines a \emph{global} problem, reaching the right place in a varying scene where centimeter errors are harmless, and a \emph{local} problem, aligning precisely where millimeters matter. Human videos suit the first, since external cameras see the whole scene and free space tolerates their embodiment error $\eta$. Sparse robot data with in-hand cameras suits the second. We therefore design a policy that separates the two sources in data, observation, and action: (1) one learns from $\mathcal{D}_h$, sees the global scene, and predicts absolute poses, (2) one learns only from $\mathcal{D}_r$, sees the in-hand view, and predicts gripper-frame corrections, (3) a gate learned from intervention labels decides which actor is in control.

\subsection{Dual Actors for Coverage and Precision} 
\label{sec:dual}

\begin{figure}[t]
    \centering
    \includegraphics[width=0.8\linewidth]{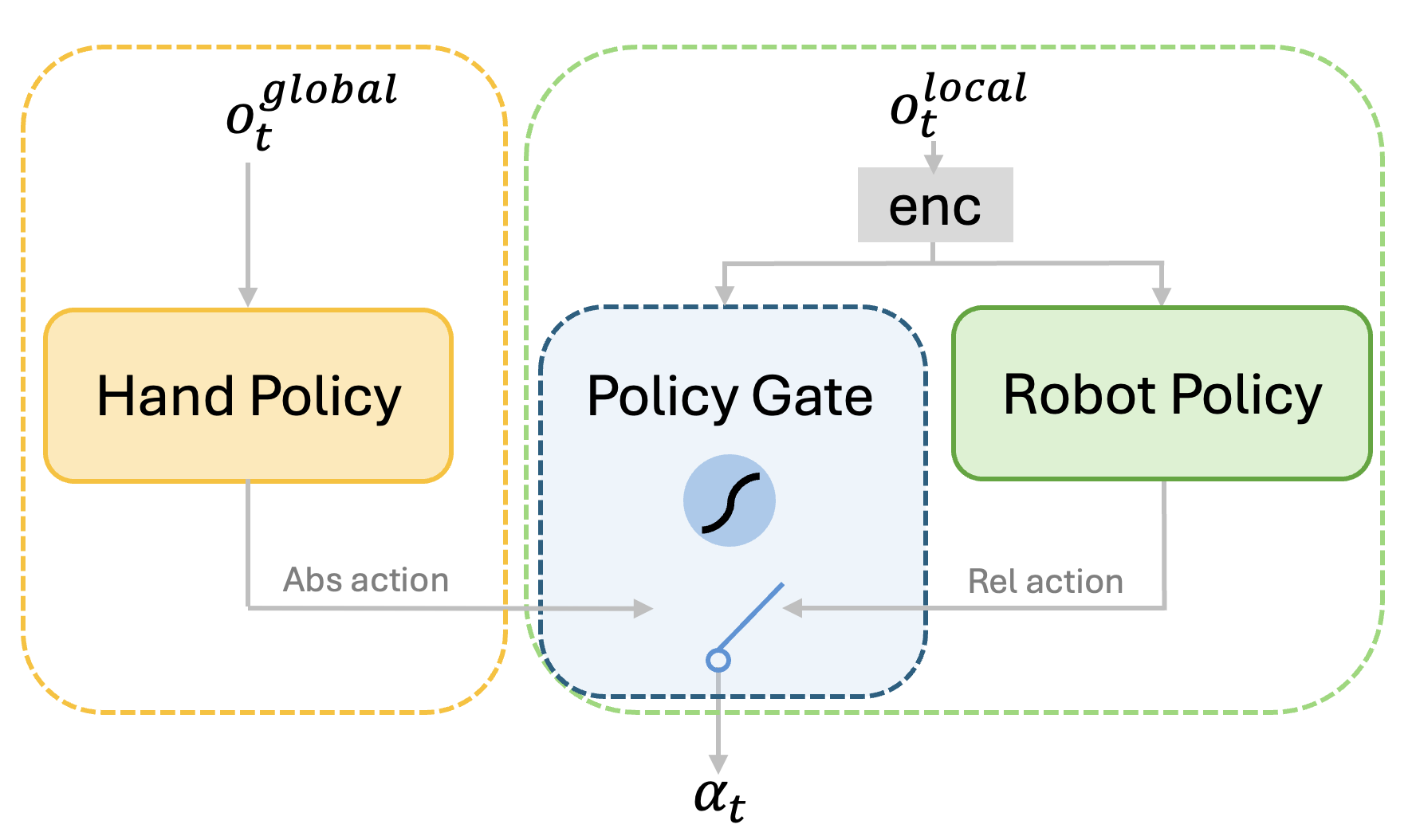}
    \caption{\textbf{Architecture.} The hand policy maps the global observation to absolute actions. The robot policy maps the in-hand view to relative actions. The policy gate shares the robot policy's encoder and switches which actor to use.}
  \label{fig:architecture}
\end{figure}

As shown in Fig.~\ref{fig:architecture}, DUO Policy consists of a hand policy $\pi_{\text{hand}}$, pretrained on human or any cross-embodiment data, and a robot policy $\pi_{\text{robot}}$, trained on sparse robot data. They observe the scene differently: the global observation $o_t^{\text{global}}$ is a scene point cloud fused from calibrated external RGB-D cameras and canonicalized by $\phi$ (Sec.~\ref{sec:canon}), available in both datasets, and the local observation $o_t^{\text{local}}$ is the RGB image from wrist-mounted in-hand cameras, available only on the robot. A state-dependent binary gate $\gamma(o_t^{\text{local}}) \in \{0,1\}$ selects which branch commands the next $K$ end-effector poses $a_{t+1:t+K} = \pi_{\text{DUO}}(o_t)$ from the current pose $T_t \in SE(3)$:
\begin{equation}
    \pi_{\text{DUO}}(o_t) =
    \begin{cases}
        T_t\,\pi_{\text{robot}}(o_t^{\text{local}}), & \text{if } \gamma(o_t^{\text{local}}) = 1, \\
        \pi_{\text{hand}}(o_t^{\text{global}}), & \text{if } \gamma(o_t^{\text{local}}) = 0.
    \end{cases}
    \label{eq:duo}
\end{equation}
$\pi_{\text{hand}}$ predicts an absolute world-frame pose and $\pi_{\text{robot}}$ a relative motion in the current gripper frame, so the gate selects between the two commanded poses rather than mixing actions.
When $\gamma=0$, the robot is in free space and follows the human prior. When $\gamma=1$, it has entered a precision-critical state and the robot policy takes over. The branches share no parameters, and the robot policy acts only when $\gamma=1$ and trains on batches that are half operator corrections, so behavior in precision-critical states is learned mainly from these corrections rather than from the human prior's error $\eta$. Moreover, each branch can use its own observation and action space.

\textbf{Global / local separation.} Robot corrections are expensive, so a correction learned at one placement should carry over to others. For example, once the robot can grasp the coffee pod at location A, it should not relearn the grasp at location B. This requires the robot actor's input and output to stay unchanged when the scene moves. Therefore, we further constrain the observation and action spaces of the two actors separately. $\pi_{\text{hand}}$ observes the global point cloud $o_t^{\text{global}}$ and predicts absolute world-frame poses, covering the workspace from dense human data. The gate and $\pi_{\text{robot}}$ observe only the in-hand view $o_t^{\text{local}}$, and $\pi_{\text{robot}}$ predicts relative motions in the current gripper frame. Neither depends on where the pod sits, so corrections transfer across placements (Proposition~\ref{prop:equiv}). Since its observations and actions share the world frame, we augment both with a common rotation $R \in SO(2)$ about the gravity axis, which makes $\pi_{\text{hand}}$ approximately equivariant to scene rotations. Because the branches share no parameters, each normalizes its actions with its own statistics, which preserves the scale of the sparse robot corrections (Appendix~\ref{appendix:extra}).

\begin{proposition}[$SE(2)$ equivariance]\label{prop:equiv}
Let $T_g \in SE(2)$ denote a planar rigid transformation of the scene about the gravity axis that acts on the state $s_t$ by transforming the scene and the gripper pose together. Let $o_t^{\text{local}} = h(s_t)$ denote the in-hand view, with $h(T_g \cdot s_t) = h(s_t)$. Writing $\pi_{\text{DUO}}(s_t)$ for $\pi_{\text{DUO}}$ applied to the observations of $s_t$, the gate, which sees $s_t$ only through $h$, is invariant to global scene transformations, and the commanded pose is globally equivariant when $\gamma=1$:
\begin{equation}
    \begin{aligned}
        \gamma\big(h(T_g \cdot s_t)\big) &= \gamma\big(h(s_t)\big), \\
        \pi_{\text{DUO}}(T_g \cdot s_t) &= T_g\, \pi_{\text{DUO}}(s_t) \quad \text{if } \gamma = 1.
    \end{aligned}
    \label{eq:equiv}
\end{equation}
When $\gamma=0$, $\pi_{\text{hand}}$ is softly $SO(2)$-equivariant through rotation augmentation (Appendix~\ref{appendix:proof}).
\end{proposition}

\begin{table*}[t]
\footnotesize
\setlength\tabcolsep{3.0pt}
\begin{center}
\caption{Simulation Results}
\label{tab:sim_result}
\begin{tabular}{@{}lccccccccccccc@{}}
\toprule
& & \multicolumn{3}{c}{Stack D1} & \multicolumn{3}{c}{Stack Three D1} & \multicolumn{3}{c}{HammerCleanup D1} & \multicolumn{3}{c}{Coffee D2} \\
\cmidrule(lr){3-5} \cmidrule(lr){6-8} \cmidrule(lr){9-11} \cmidrule(lr){12-14}
Method & Obs & 20 & 30 & 40 & 20 & 30 & 40 & 20 & 30 & 40 & 20 & 30 & 40 \\

\midrule
\textbf{DUO} & PCD & 85{\scriptsize$\pm$3} & 92{\scriptsize$\pm$3} & 94{\scriptsize$\pm$1} & \textbf{54}{\scriptsize$\pm$3} & \textbf{63}{\scriptsize$\pm$5} & \textbf{66}{\scriptsize$\pm$3} & 28{\scriptsize$\pm$1} & \textbf{37}{\scriptsize$\pm$2} & \textbf{46}{\scriptsize$\pm$4} & \textbf{61}{\scriptsize$\pm$2} & \textbf{69}{\scriptsize$\pm$6} & \textbf{71}{\scriptsize$\pm$1} \\
DP3~\citep{dp3} & PCD & 86{\scriptsize$\pm$4} & 95{\scriptsize$\pm$1} & 95{\scriptsize$\pm$2} & 8{\scriptsize$\pm$2} & 25{\scriptsize$\pm$5} & 45{\scriptsize$\pm$3} & 13{\scriptsize$\pm$3} & 14{\scriptsize$\pm$2} & 13{\scriptsize$\pm$1} & 28{\scriptsize$\pm$4} & 52{\scriptsize$\pm$1} & 51{\scriptsize$\pm$4} \\
DP3-rel & PCD & \textbf{95}{\scriptsize$\pm$1} & \textbf{97}{\scriptsize$\pm$2} & \textbf{98}{\scriptsize$\pm$2} & 36{\scriptsize$\pm$6} & 34{\scriptsize$\pm$3} & 38{\scriptsize$\pm$3} & 25{\scriptsize$\pm$4} & 30{\scriptsize$\pm$5} & 37{\scriptsize$\pm$3} & 26{\scriptsize$\pm$8} & 37{\scriptsize$\pm$6} & 33{\scriptsize$\pm$1} \\
DiffPo~\citep{diffpo} & RGB & 79{\scriptsize$\pm$0} & 77{\scriptsize$\pm$2} & 84{\scriptsize$\pm$2} & 35{\scriptsize$\pm$3} & 43{\scriptsize$\pm$2} & 39{\scriptsize$\pm$2} & \textbf{33}{\scriptsize$\pm$1} & 33{\scriptsize$\pm$4} & 38{\scriptsize$\pm$1} & 41{\scriptsize$\pm$2} & 53{\scriptsize$\pm$4} & 56{\scriptsize$\pm$4} \\
Phantom~\cite{phantom} & RGB & \multicolumn{3}{c}{--------- 13{\scriptsize$\pm$2} ---------} & \multicolumn{3}{c}{--------- 0{\scriptsize$\pm$0} ---------} & \multicolumn{3}{c}{--------- 2{\scriptsize$\pm$2} ---------} & \multicolumn{3}{c}{--------- 0{\scriptsize$\pm$0} ---------} \\
Phantom-SFT~\cite{phantom} & RGB & 11{\scriptsize$\pm$0} & 15{\scriptsize$\pm$1} & 26{\scriptsize$\pm$2} & 0{\scriptsize$\pm$0} & 0{\scriptsize$\pm$0} & 0{\scriptsize$\pm$0} & 19{\scriptsize$\pm$0} & 6{\scriptsize$\pm$2} & 12{\scriptsize$\pm$3} & 5{\scriptsize$\pm$0} & 6{\scriptsize$\pm$1} & 5{\scriptsize$\pm$2} \\
Masquerade~\cite{masquerade} & RGB & 17{\scriptsize$\pm$1} & 32{\scriptsize$\pm$2} & 47{\scriptsize$\pm$2} & 1{\scriptsize$\pm$1} & 1{\scriptsize$\pm$1} & 5{\scriptsize$\pm$2} & 12{\scriptsize$\pm$1} & 16{\scriptsize$\pm$2} & 22{\scriptsize$\pm$0} & 4{\scriptsize$\pm$1} & 16{\scriptsize$\pm$3} & 30{\scriptsize$\pm$2} \\
\bottomrule
\end{tabular}
\par\vspace{2pt}\parbox{\textwidth}{\footnotesize Success rates (\%) after pretraining on 300 noisy \textcolor{orange}{Panda demonstrations} (500 for HammerCleanup) and finetuning on 20, 30, or 40 \textcolor{robotgreen}{UR5 intervention episodes}, which each method collects with its own pretrained policy (Masquerade, whose pretraining predicts only keypoints and cannot act, uses the episodes collected by Phantom). Each policy is evaluated at its final checkpoint (epoch 600) on 100 unseen initial configurations. Results are over 3 finetuning seeds. Values are mean $\pm$ std; $\pm$0 indicates a standard deviation below 0.5 points; averages in the text are computed from unrounded means. Bold indicates the best. Obs: pretraining observation (PCD: point cloud from five cameras, RGB: front-view image). All methods also receive in-hand RGB images during finetuning.}
\end{center}

\end{table*}

The gate's invariance is what makes the equivariance hold. If $\gamma(h(s_t))=1$, then
\begin{equation}
    \begin{aligned}
        \pi_{\text{DUO}}(T_g \cdot s_t)
        &\overset{(a)}{=} (T_g T_t)\, \pi_{\text{robot}}\big(h(T_g \cdot s_t)\big) \\
        &\overset{(b)}{=} T_g\,\big(T_t\, \pi_{\text{robot}}(h(s_t))\big)
        = T_g\, \pi_{\text{DUO}}(s_t),
    \end{aligned}
    \label{eq:gate_role}
\end{equation}
where (a) holds because $\gamma(h(T_g \cdot s_t)) = \gamma(h(s_t)) = 1$, so the transformed state is still handled by $\pi_{\text{robot}}$ at the transformed pose $T_g T_t$, and (b) because the in-hand view is unchanged. A gate that saw the global scene could switch to $\pi_{\text{hand}}$ in step (a) and break the identity. The invariance also lets intervention labels from a few placements transfer to new ones, since whether a state is precision-critical depends on the geometry around the gripper rather than on its position in the workspace (Sec.~\ref{sec:gate}). Proposition~\ref{prop:equiv} explains why sparse robot data can suffice: in precision-critical states, the robot actor's input and output do not change with the placement, so its data is spent on local geometry (e.g., pod insertion) rather than on global spatial variation. In free space, $\pi_{\text{hand}}$ is equivariant approximately, through $SO(2)$ augmentation, and absorbs workspace variation from dense human data. Giving the robot actor global observations or absolute actions breaks the proposition (Table~\ref{tab:globallocal}).

\subsection{Interventions as Labels for Precision-Critical States}
\label{sec:interv}

The gate must recognize $\mathcal{S}_{\text{crit}}$, but the tolerance $\varepsilon(s)$ that defines it is unknown. We obtain labels from $\pi_{\text{hand}}$ itself. This design choice follows our observation that cross-embodiment policies usually fail at fine-grained actions (Fig.~\ref{fig:why}), so the states where $\pi_{\text{hand}}$ needs help are the precision-critical ones. We deploy $\pi_{\text{hand}}$ on the robot in a single round, and an operator takes over whenever it is about to fail, i.e., when its error exceeds what the state tolerates. Labeling each timestep with $i_t=1$ under operator control and $i_t=0$ otherwise gives $i_t \approx \mathbbm{1}[s_t \in \mathcal{S}_{\text{crit}}]$ at no extra labeling cost, and concentrates the operator's effort where precision is required. We train $\pi_{\text{robot}}$, a separate diffusion policy, on all timesteps of $\mathcal{D}_r$, sampling intervention and non-intervention timesteps 1:1 in each batch so that operator corrections make up half of every batch, similar in spirit to the weighted behavioral cloning~\citep{liu2025robot}. Non-intervention timesteps carry $\pi_{\text{hand}}$'s own actions, but $\pi_{\text{robot}}$ is executed only when the gate selects it, i.e., in precision-critical states, where its targets are operator corrections. We also train a gate network $p_\theta$ to predict $i_t$ and set $\gamma = \mathbbm{1}[p_\theta(o_t^{\text{local}}) > 0.5]$. $p_\theta$ and $\pi_{\text{robot}}$ share an in-hand encoder initialized with pre-trained ResNet-18~\citep{he2016deep} weights:
\begin{equation}
\mathcal{L}_{\text{finetune}} = \mathcal{L}_{\text{diff}}\left(\pi_{\text{robot}}; \mathcal{D}_r\right) + \lambda\, \mathcal{L}_{\text{BCE}}\!\left(p_\theta(o_t^{\text{local}}), i_t\right),
\end{equation}
where $\lambda$ balances the two losses. The learned gate agrees with a privileged phase label (F1 0.85) and matches its task success (Section~\ref{sec:gate}).

\subsection{Canonical Observations for Cross-Embodiment Transfer}
\label{sec:canon}

$\pi_{\text{hand}}$ is trained on human observations but deployed on the robot. Similar to Phantom~\citep{phantom}, which inpaints the human and overlays a robot arm in RGB images, we reduce this mismatch with a canonicalization map $\phi$, but in 3D: $\phi$ removes whichever embodiment acts in the scene from the fused world-frame point cloud and renders the target gripper geometry at its end-effector pose, so the canonical observation is consistent across camera views. Human and robot observations thus share one space: $\pi_{\text{hand}}$ is trained on $\mathcal{D}_h = \{\tau_i^h\}_{i=1}^{N_h}$ with $\tau_i^h = \{(\phi(o^h_t), \hat{a}_{t+1:t+K})\}_t$, where $\hat{a}_{t+1:t+K}$ is the chunk of the next $K$ human hand poses retargeted to gripper poses, and deployed on $\phi(o^r_t)$. Because $\phi$ only needs the end-effector pose of the acting embodiment, it applies to any source embodiment whose end-effector pose can be estimated. We parameterize $\pi_{\text{hand}}$ as a conditional diffusion policy~\cite{diffpo} over the global point cloud, trained with the standard denoising objective.

\textbf{Human Data Canonicalization.} From three calibrated RGB-D cameras, we estimate hand poses with WiLoR~\citep{potamias2025wilor}, discard low-confidence detections, correct the translation with the depth image following~\citep{phantom}, and smooth the trajectories with a Savitzky--Golay filter. To retarget, we transform the hand poses into the world frame, ${}^W T_H = {}^W T_C \, {}^C T_H$, compute a fixed hand-to-gripper offset from the first frame that aligns the fingertips with the initial gripper pose, ${}^H T_G = ({}^W T_{H,0})^{-1} \, {}^W T_G^{\mathrm{init}}$, obtain the retargeted gripper pose $\hat{T}_t = {}^W T_{H,t} \, {}^H T_G$, and take the next $K$ poses as the action label, $\hat{a}_{t+1:t+K} = (\hat{T}_{t+1}, \dots, \hat{T}_{t+K})$. The binary gripper command is obtained by thresholding the thumb--index fingertip distance. For the observations, SAM3~\citep{carion2025sam3} segments the human in each view with the prompt ``human''. The remaining pixels are back-projected into a human-free point cloud, the gripper is rendered at the current pose $\hat{T}_t$, and the result is cropped to the workspace and downsampled to 1024 points. See Figure~\ref{fig:pipeline} in Appendix~\ref{appendix:human} and Fig.~\ref{fig:hand_pcd} for the pipeline and examples.

\textbf{Robot Data Canonicalization.} At deployment, we reconstruct a world-frame point cloud from the same cameras, compute the robot geometry by forward kinematics on its URDF, and remove observed points within 4\,cm of the posed robot meshes using a KD-tree query. The finger links are excluded from this removal so that objects near the contact region are kept. We then add a synthetic gripper point cloud at the measured end-effector pose and gripper state. This runs at roughly 33\,Hz. Without the $\phi$ alignment, success drops by 31.7 percentage points (Section~\ref{sec:ablation}).

\begin{figure*}[t] 
    \centering 
    \setlength{\tabcolsep}{2pt} 
    \renewcommand{\arraystretch}{0} 
    \begin{tabular}{cccccc} 
        \begin{overpic}[width=0.155\linewidth]{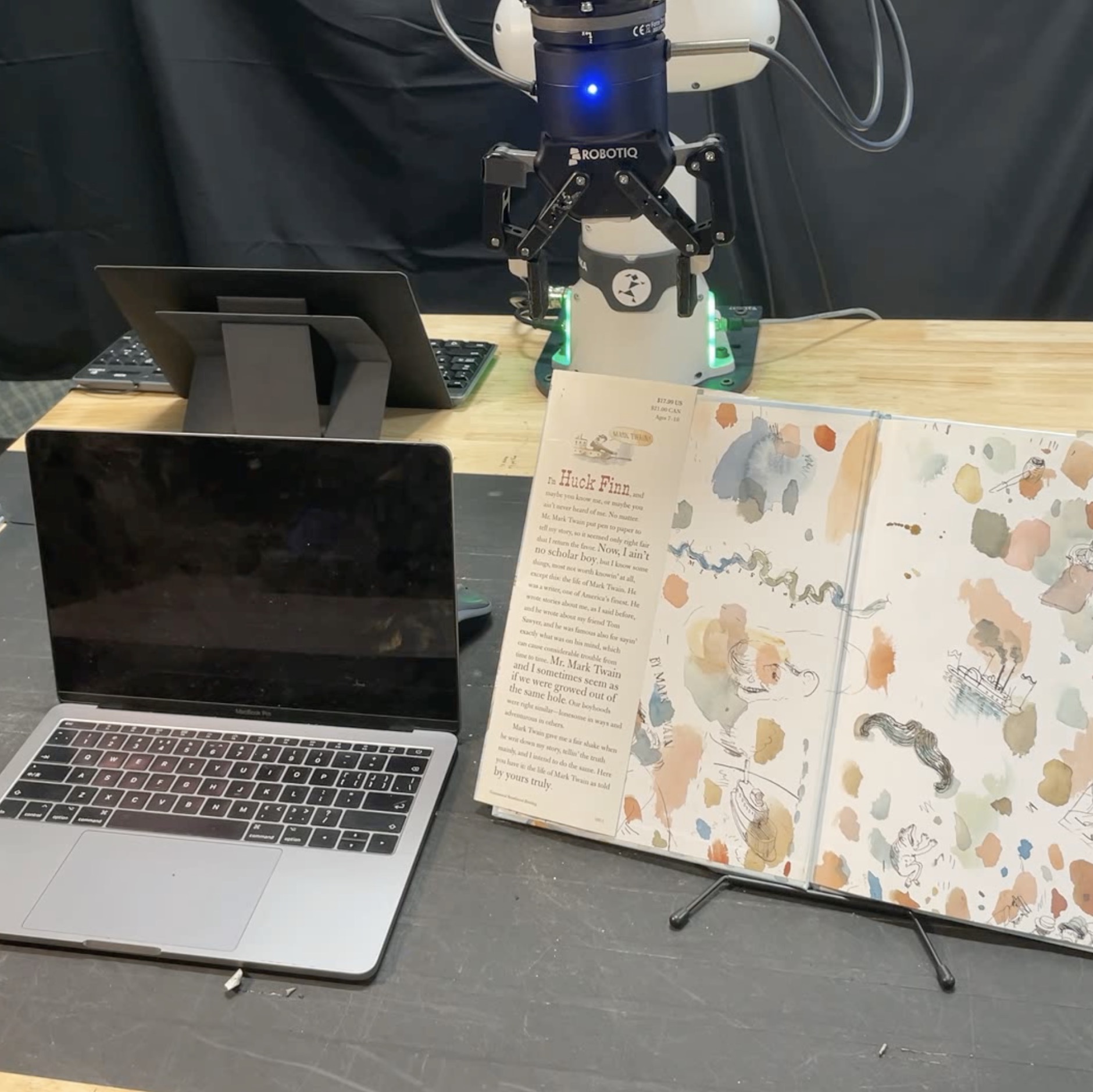}
            \put(1,85){
                \textcolor{white}{\small \textbf{(a)}}
            }
        \end{overpic} &
        \includegraphics[width=0.155\linewidth]{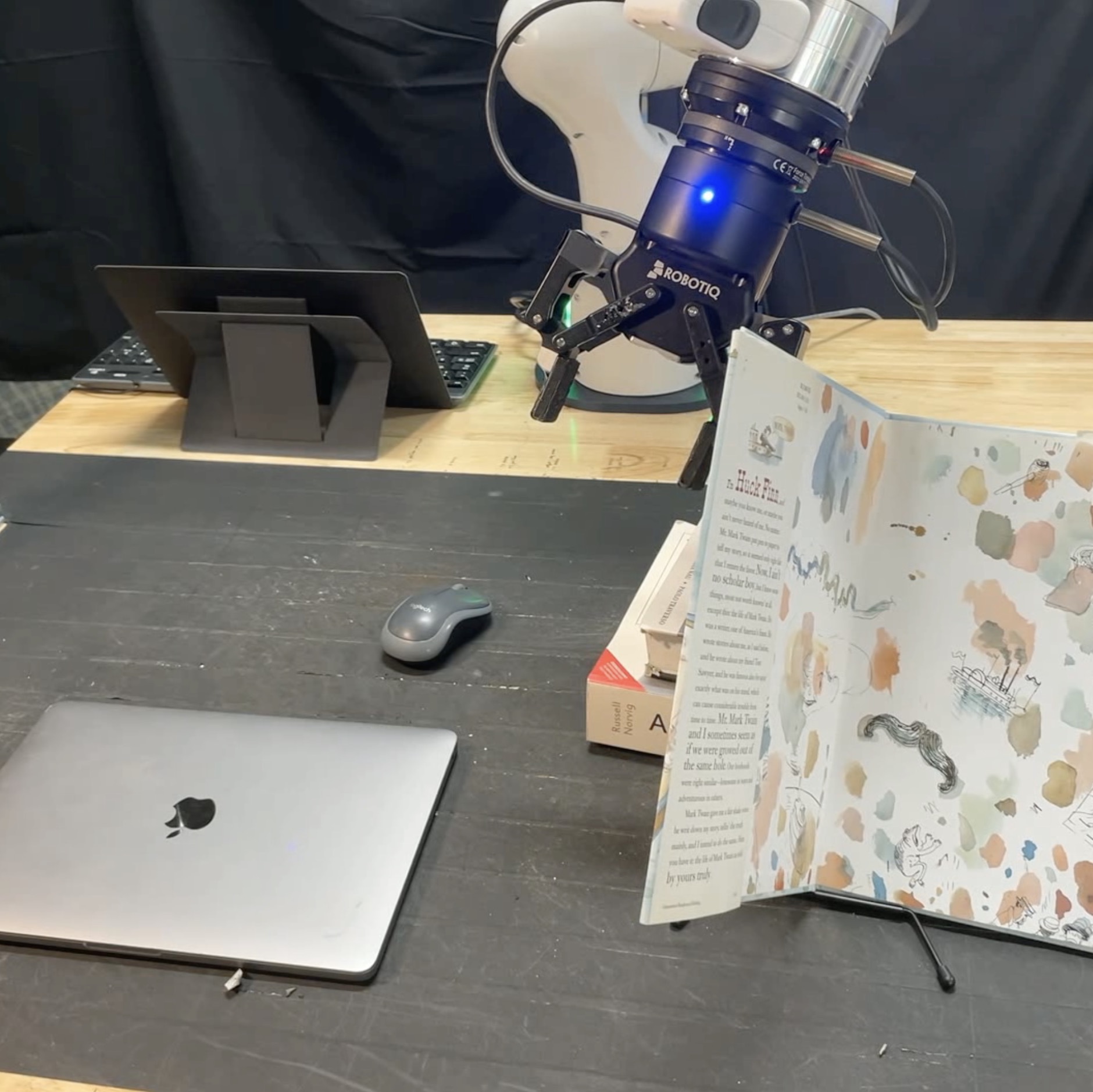} & 
        \includegraphics[width=0.155\linewidth]{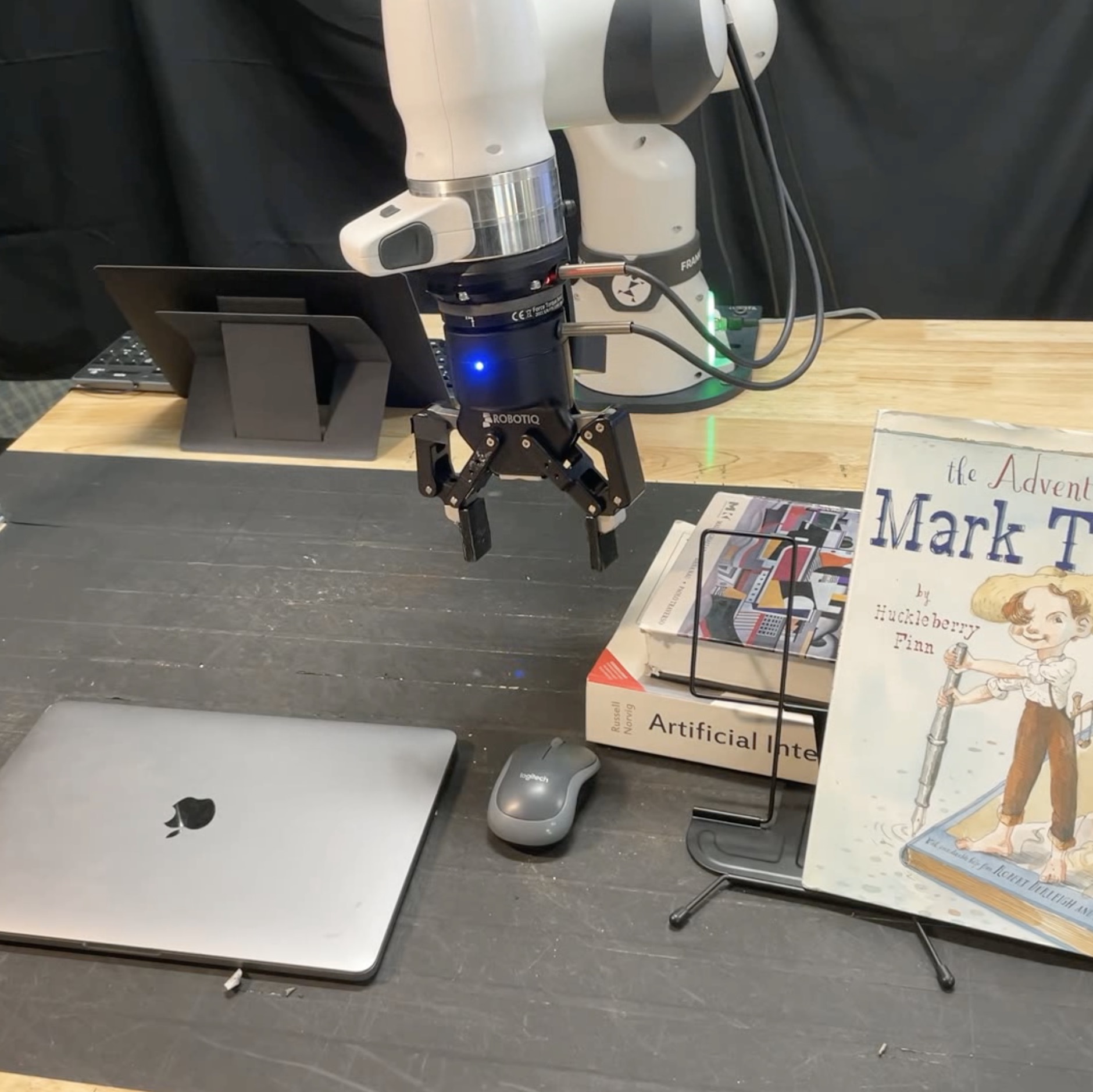} & 
        \begin{overpic}[width=0.155\linewidth]{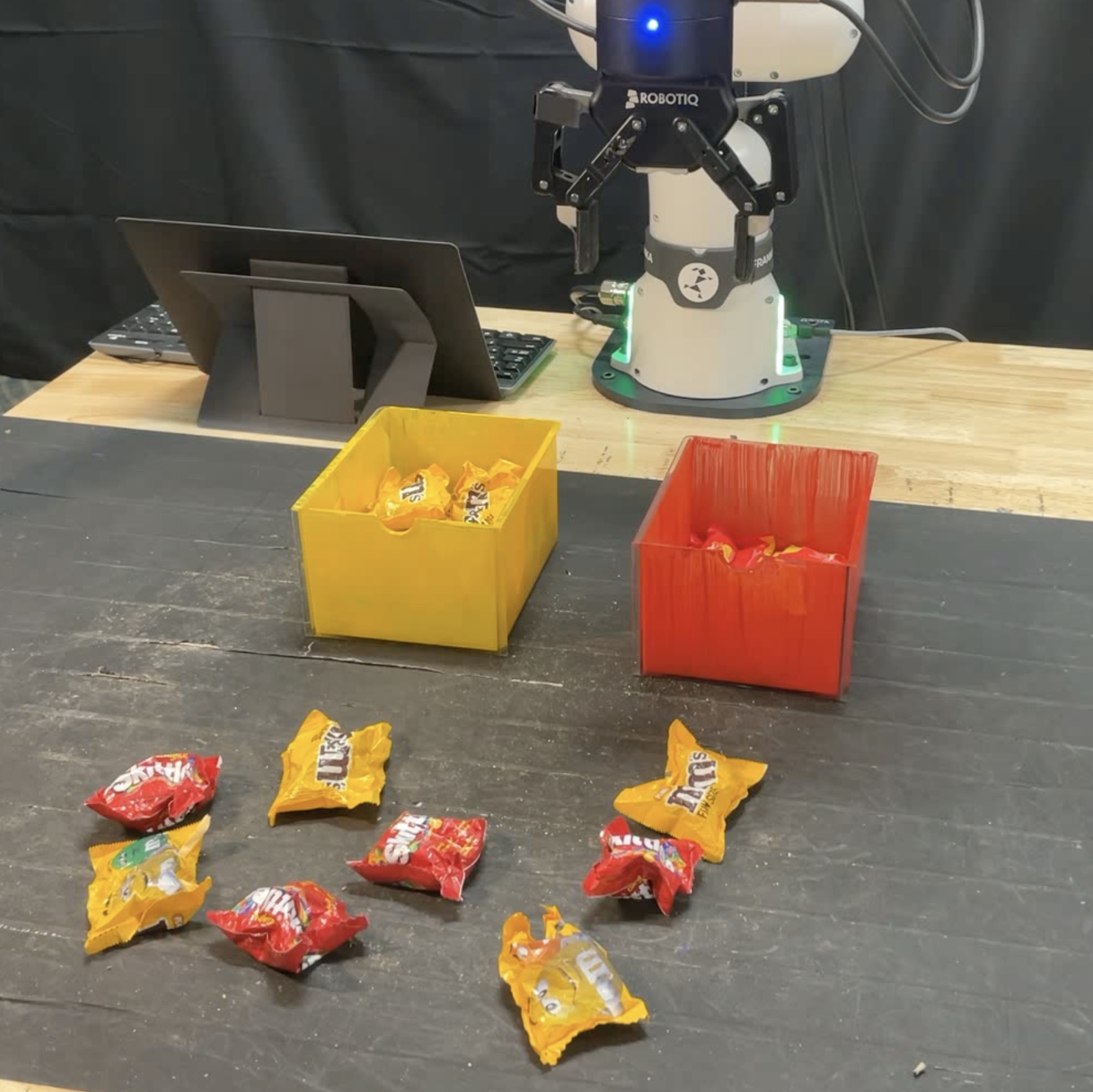}
            \put(1,85){
                \textcolor{white}{\small \textbf{(b)}}
            }
        \end{overpic} &
        \includegraphics[width=0.155\linewidth]{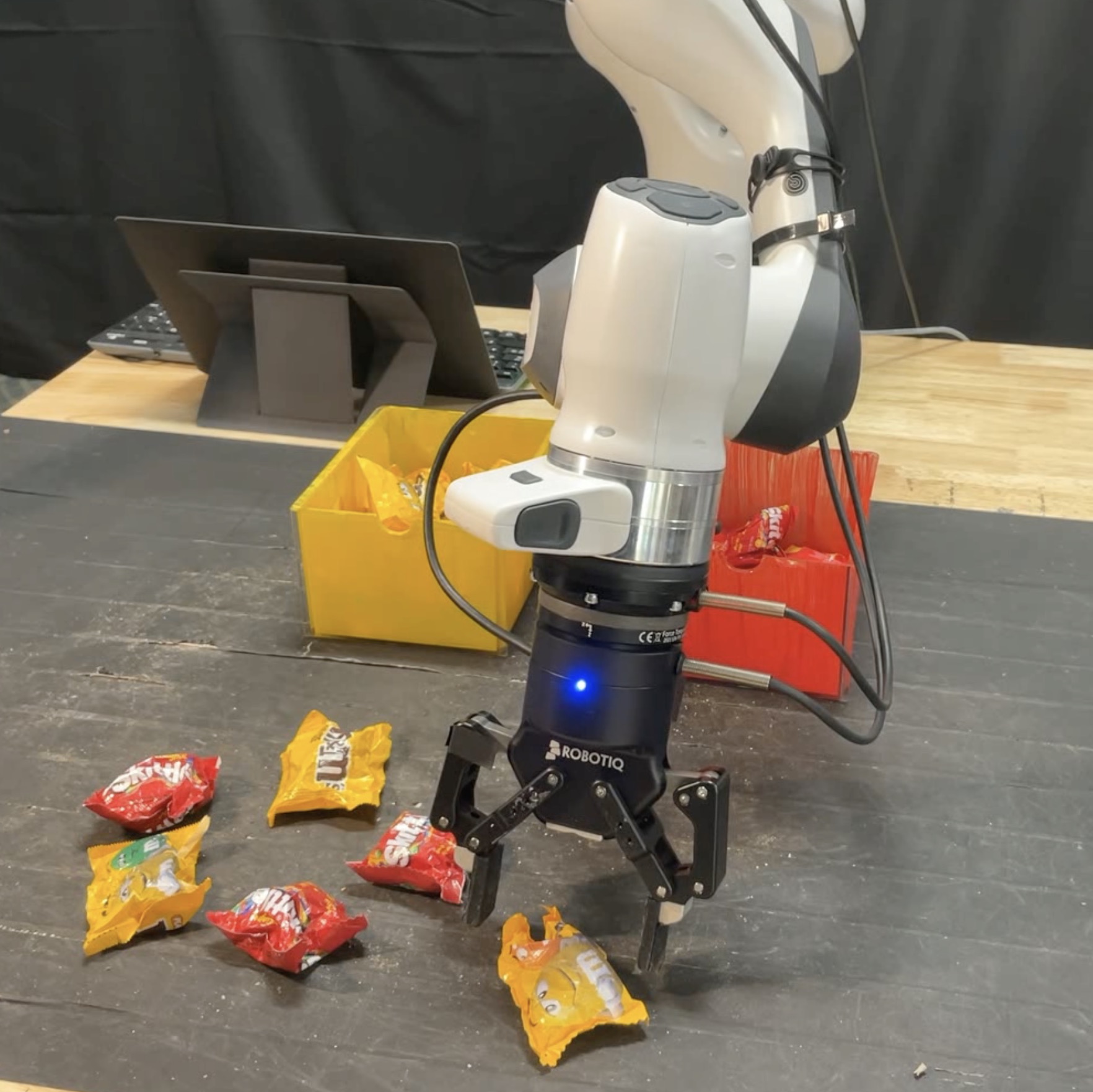} & 
        \includegraphics[width=0.155\linewidth]{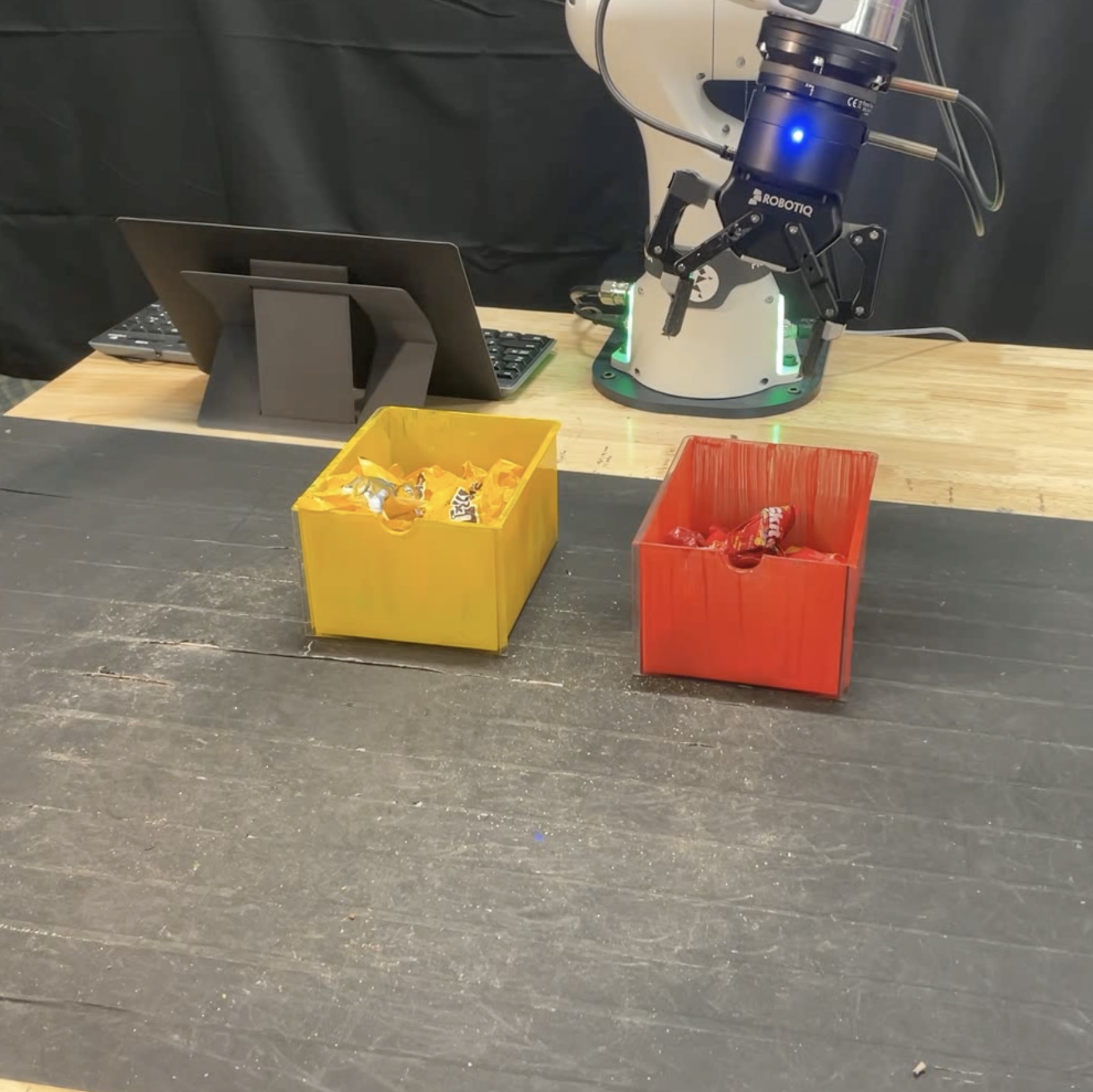} \\[4 pt] 
    \end{tabular} 
    
    \begin{tabular}{cccccc} 
        \begin{overpic}[width=0.155\linewidth]{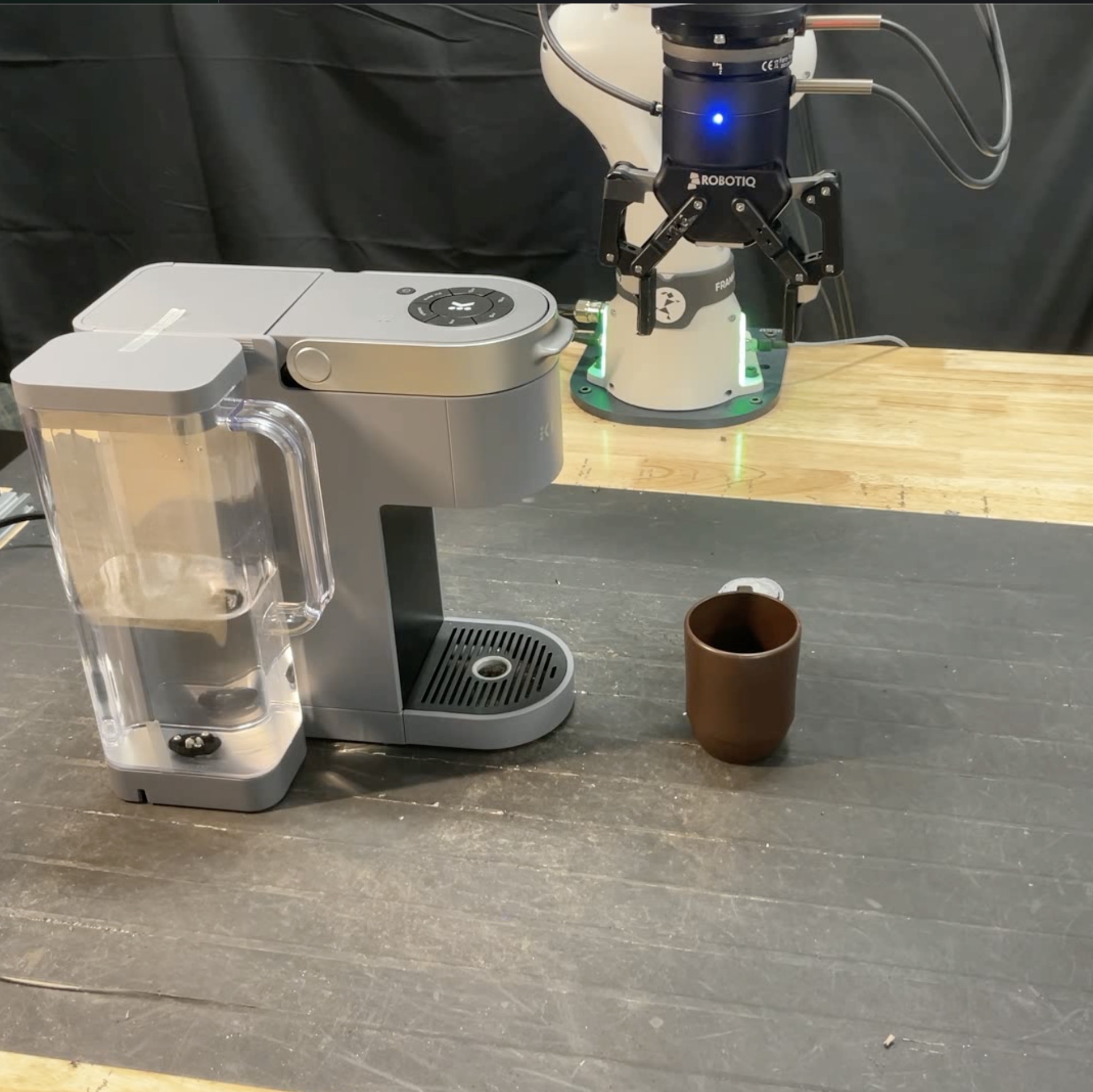}
            \put(1,85){
                \textcolor{white}{\small \textbf{(c)}}
            }
        \end{overpic} &
        \includegraphics[width=0.155\linewidth]{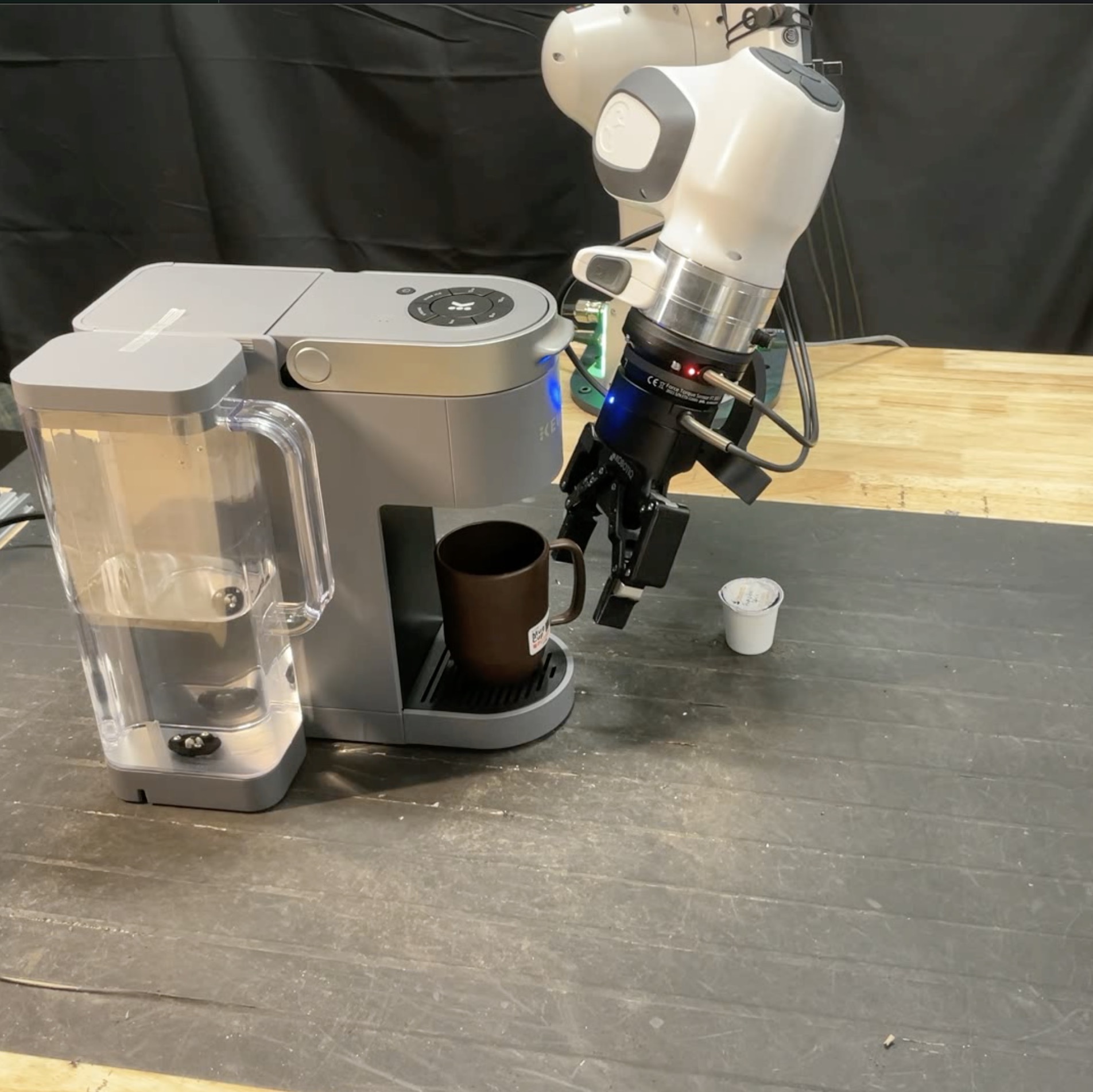} & 
        \includegraphics[width=0.155\linewidth]{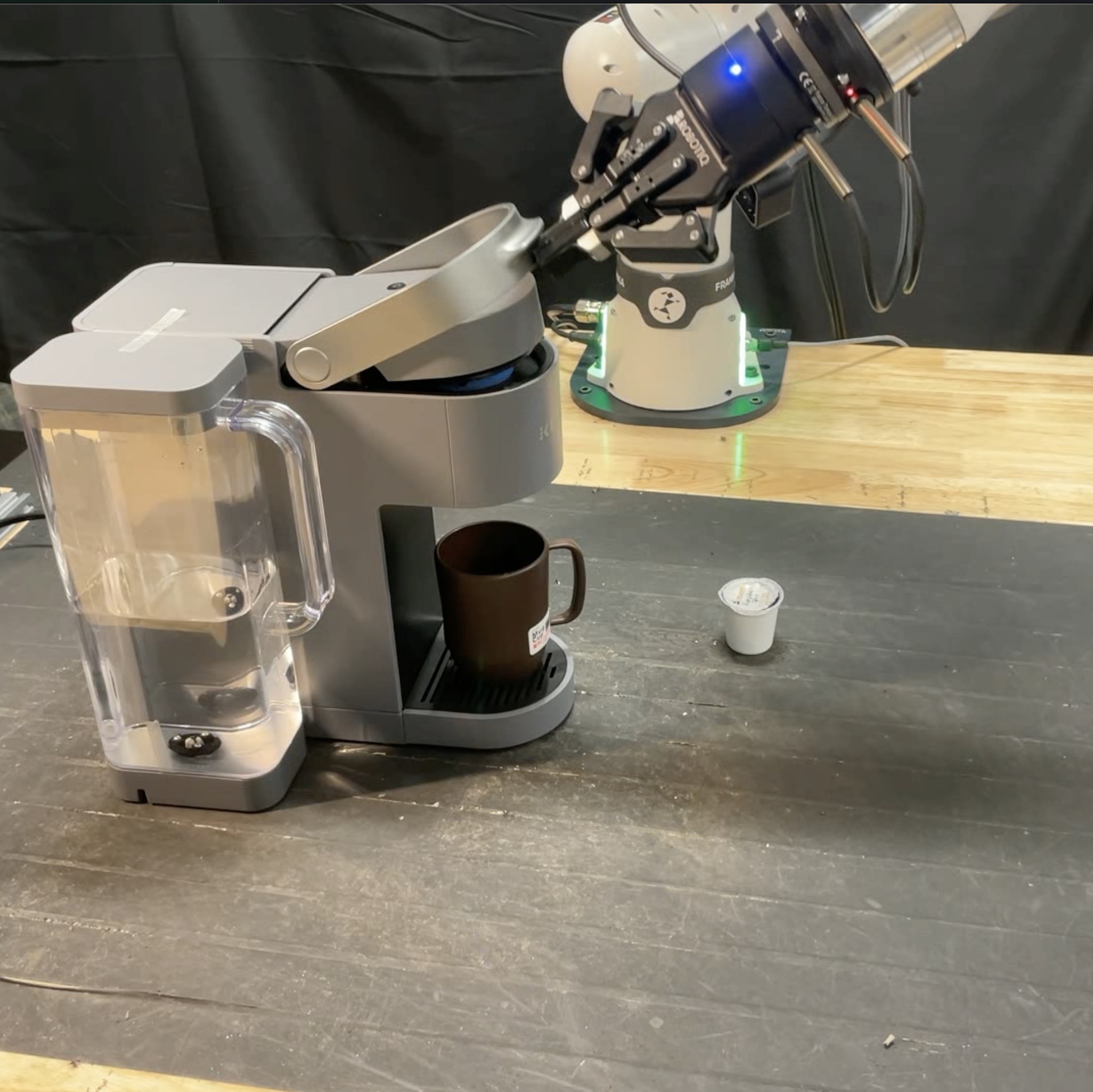} & 
        \includegraphics[width=0.155\linewidth]{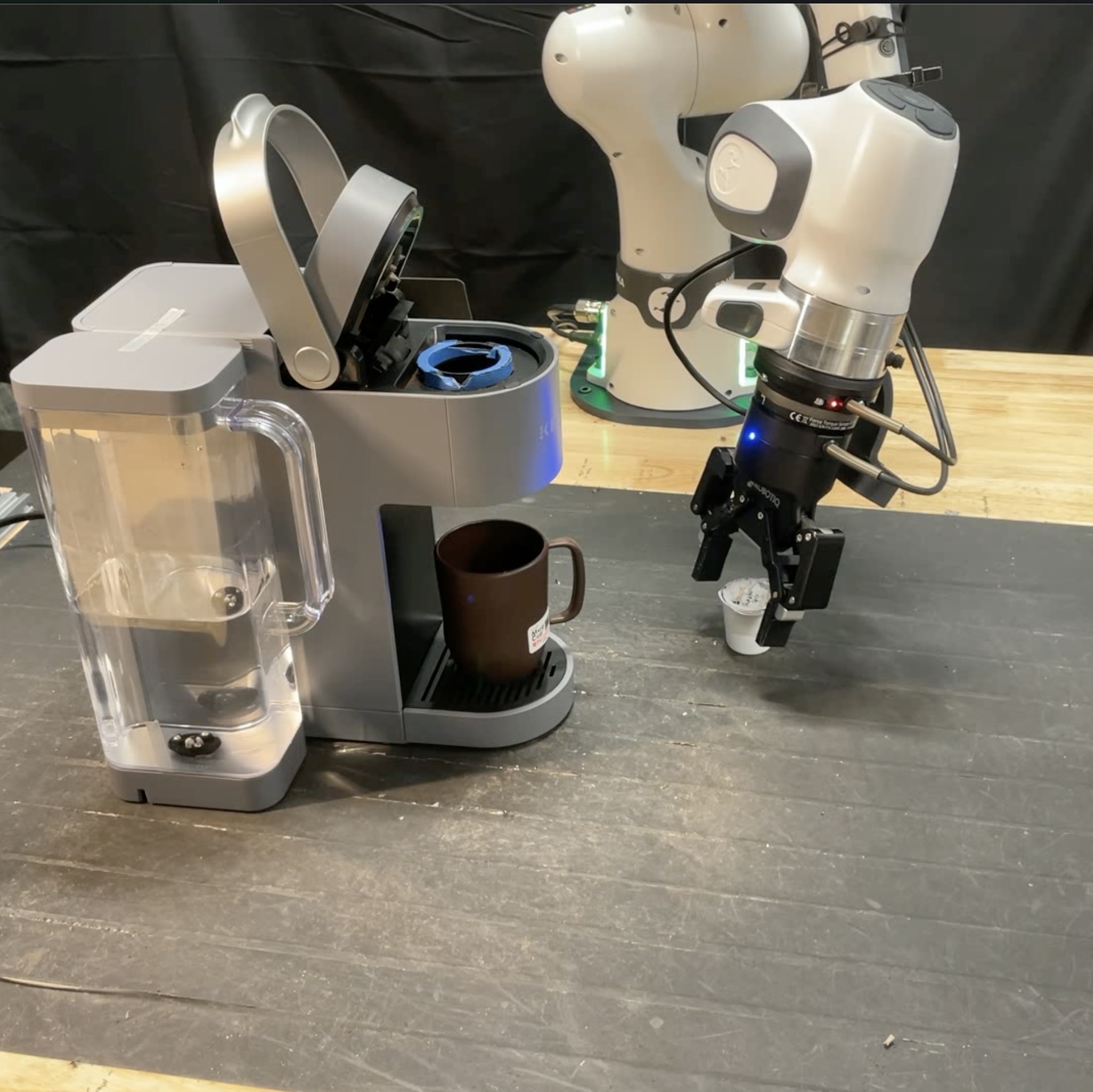} & 
        \includegraphics[width=0.155\linewidth]{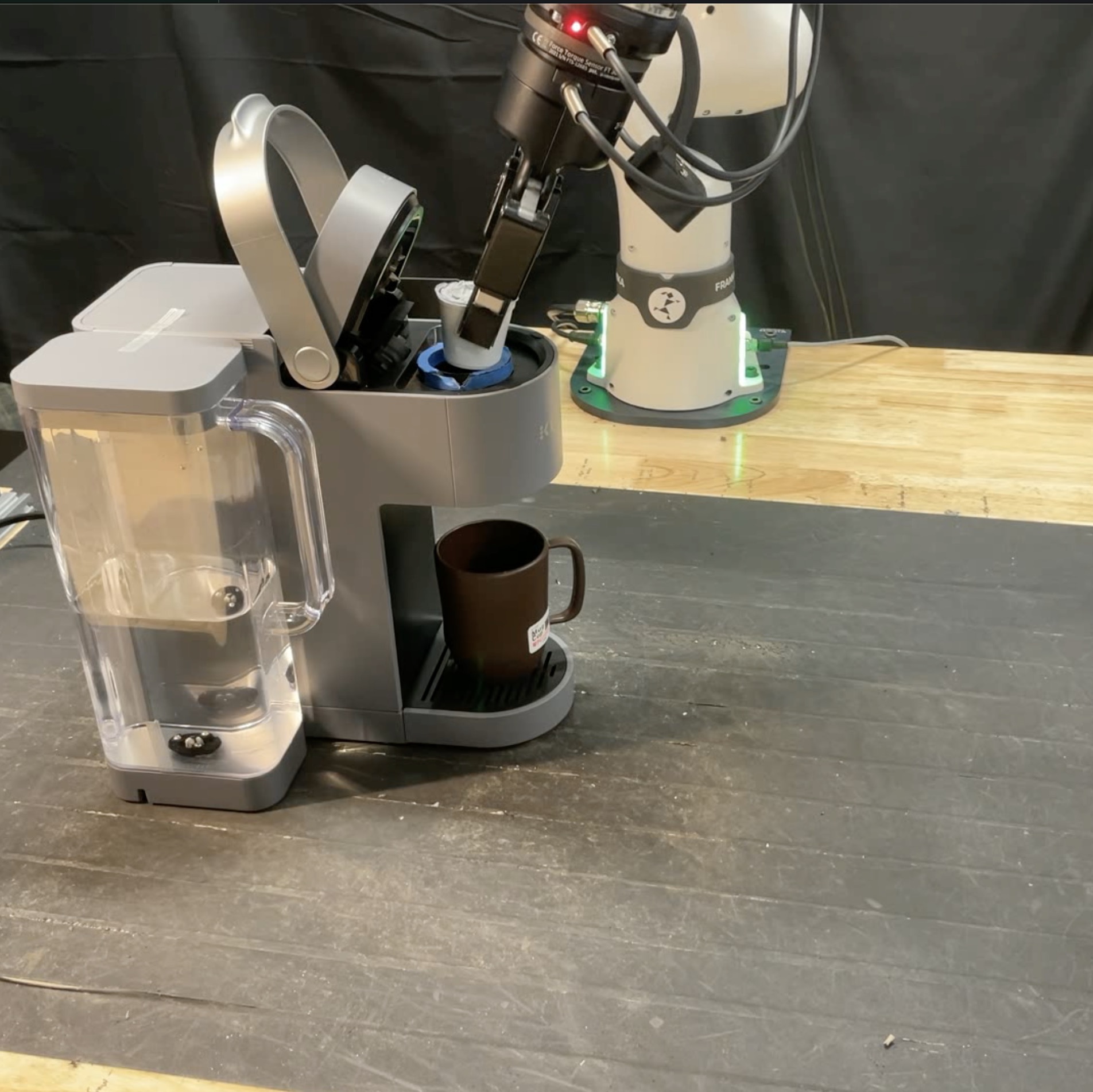} & 
        \includegraphics[width=0.155\linewidth]{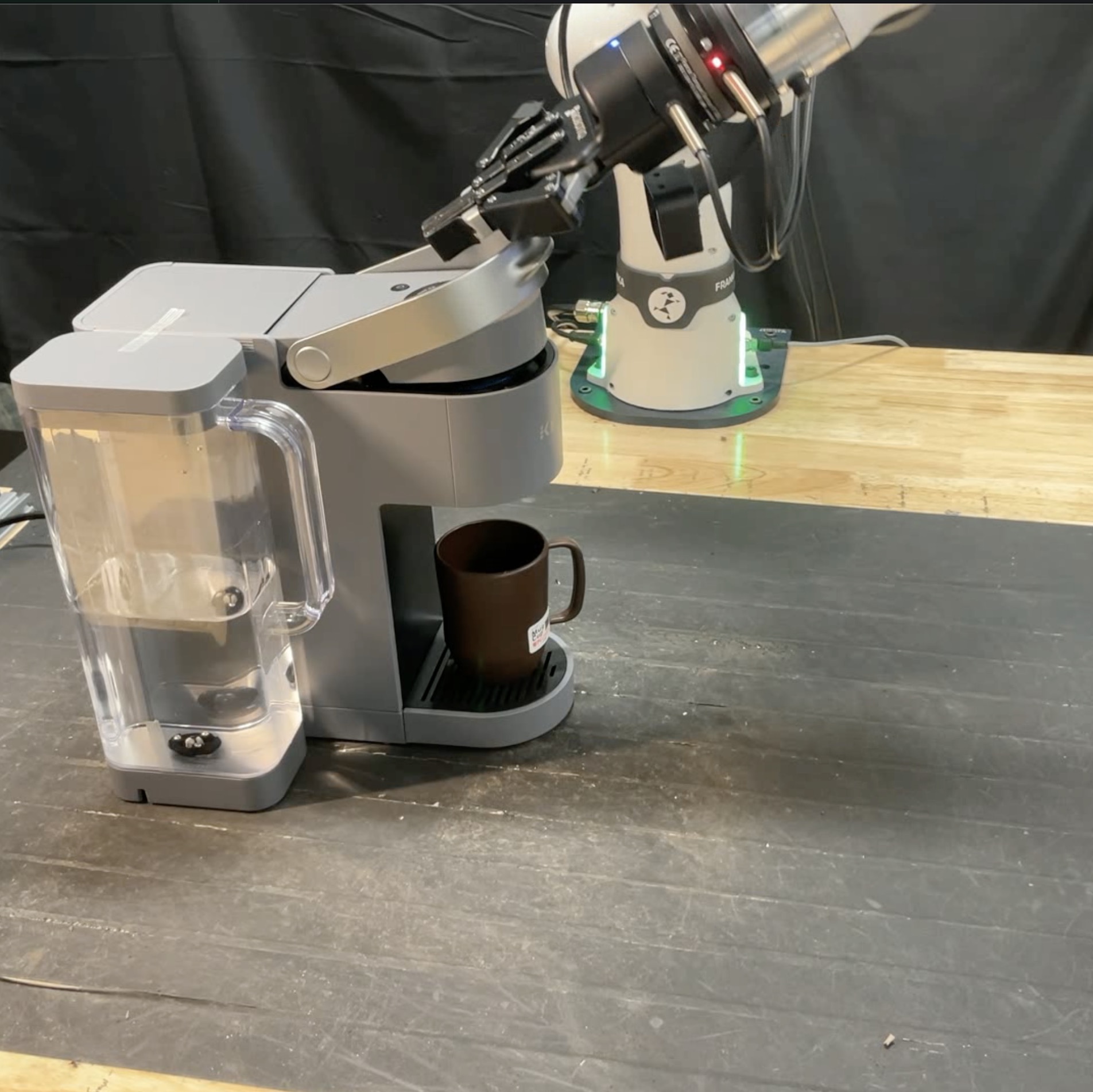} \\[4 pt] 
    \end{tabular} 
    
    \caption{\textbf{Real-world Tasks.}
        \textbf{(a) Desk Cleanup:} close the laptop and the book, then place the mouse aligned with the laptop.
        \textbf{(b) Candy Sorting:} sort cluttered candy bags into boxes by color.
        \textbf{(c) K-Cup Coffee Making:} place the mug, open the machine, insert the pod, close the machine, and press brew.
    }
    \label{fig:tasks} 
\end{figure*}

\section{Experiments}

We evaluate DUO Policy in simulation and the real world to answer four questions: \textbf{(Q1)} Does DUO outperform prior cross-embodiment methods? \textbf{(Q2)} Where does each data source help? \textbf{(Q3)} Does the gate find precision-critical states? \textbf{(Q4)} Which components matter?

\textbf{Simulation Setup.} In simulation, we use four MimicGen~\citep{mandlekar2023mimicgen} tasks. All methods pretrain on the same noisy Panda demonstrations, which, like human videos, lack in-hand views and carry embodiment gaps (simulated by a pose offset drawn uniformly within $\pm$2\,cm per episode). Each method then collects its own UR5 data: a scripted MimicGen intervener, which decides whether to take over after each executed 8-step action chunk, takes over its pretrained policy at precision-critical states, detected by a privileged phase oracle, so all methods receive the same number of successful robot episodes, each collected on its own state distribution. Phantom~\citep{phantom} trains zero-shot on inpainted human demonstrations, and Phantom-SFT finetunes it on robot data. Masquerade~\citep{masquerade} pretrains to predict keypoints rather than actions, then co-trains robot actions with keypoint prediction as an auxiliary task. Since its pretrained model cannot act, it uses the episodes collected with Phantom. We also compare to Diffusion Policy~\citep{diffpo} and DP3~\citep{dp3}, both finetuned with SFT. All baselines use absolute actions except DP3-rel. Every method receives in-hand images during finetuning. The ablations (Table~\ref{tab:ablation}) follow the protocol of Table~\ref{tab:sim_result}. The other analyses use their own robot data and evaluation sets (e.g., separate seen and unseen placements for Table~\ref{tab:gate}), so success rates are comparable only within a table or figure.

\textbf{Real-world Setup.} In the real world, we evaluate three tasks (Fig.~\ref{fig:tasks}) and compare against DP3 with in-hand images and Masquerade-3D, which co-trains on 3D human and robot demonstrations and shares our canonicalization $\phi$, rendering the robot gripper in place of the human points. All methods receive the same data-collection time budget. Baselines use teleoperated demonstrations instead of DAgger, which in preliminary trials gave them higher success within this budget, since their first-round policies cannot yet complete the approach reliably enough to intervene on. The processing pipeline and hyperparameters are shared across all tasks, with no per-task tuning. Human demonstrations are collected with a real-time interface that runs WiLoR online, retargets the hand to a gripper pose, and solves inverse kinematics, alerting the demonstrator whenever a motion is infeasible for the robot. 

\textbf{Implementation details.} All policies use a prediction horizon of $K=16$ and an execution horizon of 8, with an observation horizon of 1 for $\pi_{\text{hand}}$ and 2 for $\pi_{\text{robot}}$. Appendix~\ref{appendix:impl} lists the full settings. We train with batch size 256, learning rate $10^{-4}$ with a cosine schedule and 500 warmup steps, and weight decay $10^{-6}$, for 400 pretraining and 600 finetuning epochs.

\subsection{Does DUO Outperform Prior Methods?}
\label{sec:compare}

As shown in Table~\ref{tab:sim_result}, our method outperforms all baselines in 8 of the 12 task. The exceptions are Stack D1 at every budget and HammerCleanup with 20 episodes. Stack D1 is the control case our hypothesis predicts: its loose precision requirement makes the embodiment gap less harmful, so single policies trained on the noisy data already succeed (up to 98\%) and leave little for the robot actor to correct. Where precision matters, on Stack Three D1 and Coffee D2, whose stacking and insertion steps the noisy pretraining data cannot resolve, DUO leads the best baseline by 15--21 points at every budget. DUO's advantage thus grows with the precision a task demands. Averaged over all 12 settings, DUO exceeds the baseline with the highest average, DiffPo, by 12.9 percentage points (63.8\% vs.\ 50.9\%). Against the best baseline in each setting separately, the average margin is 8.3 points. DP3-rel averages 48.6\% and reaches 33\% on Coffee D2 (vs.\ 71\%), which shows that DUO's gains come from the dual-actor architecture, not from the action frame alone.

\begin{figure}[t]
  \centering
  \includegraphics[width=\linewidth]{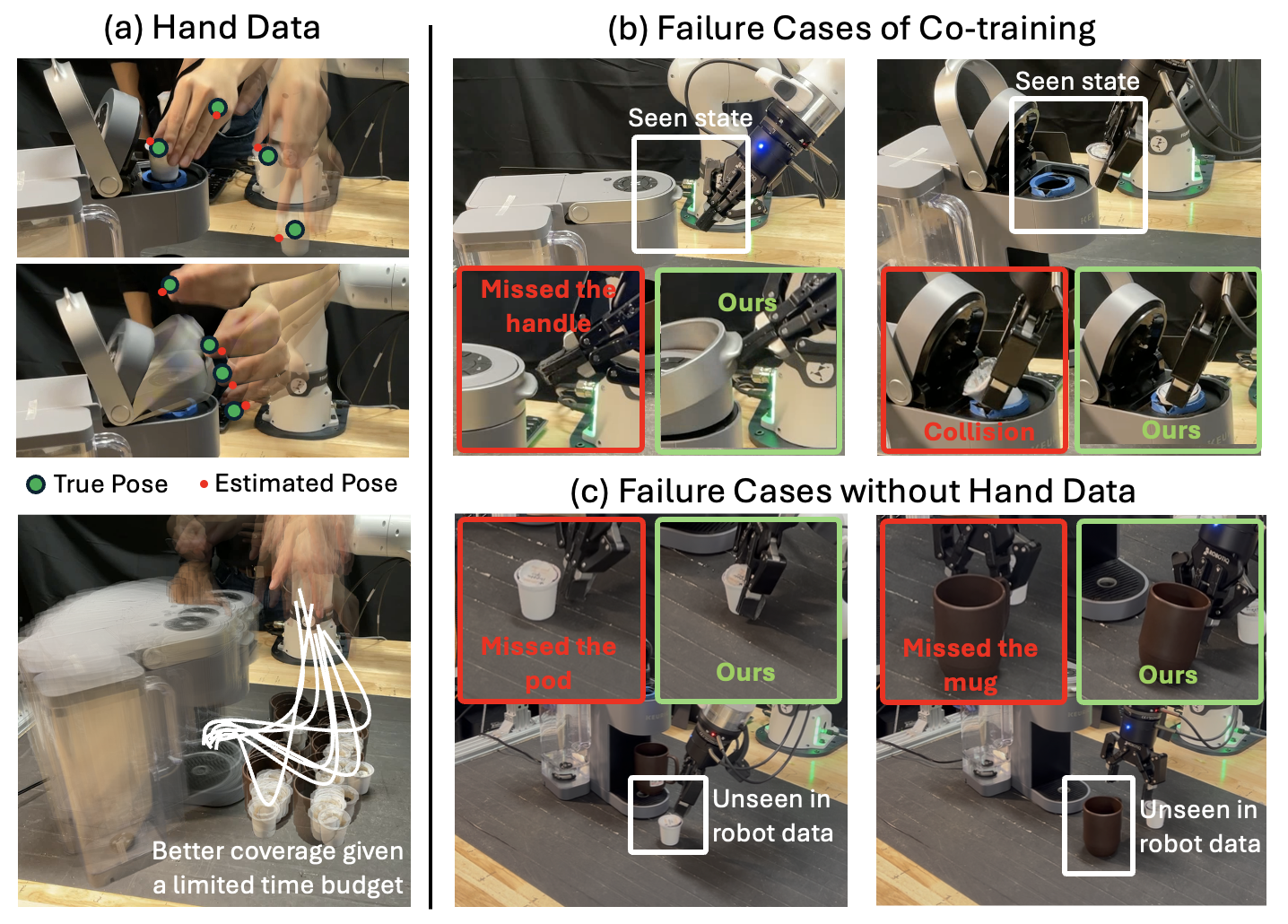}
  \caption{\textbf{Real-World Baseline Failure Cases.} (a) Hand data: estimated poses (red) deviate from the true poses (green), but a limited time budget still covers the workspace broadly. (b) Co-training on hand data (Masquerade-3D) fails at precision-critical steps. (c) Training without hand data (DP3) fails at poses unseen in the robot data. DUO (green) succeeds in all four cases.}
  \label{fig:case_study}
\end{figure}

\begin{table}[t]
  \centering
  \caption{Real-World Task Success Rates}
  \label{tab:real_result}
  \footnotesize
  \setlength\tabcolsep{4pt}
  \begin{tabular}{l c c c}
  \toprule
  \textbf{Method} & \textbf{Desk Cleanup} & \textbf{Candy Sort} & \textbf{K-Cup Coffee} \\
  \midrule
  \textbf{DUO} & \textbf{17/20} & \textbf{72/80} & \textbf{14/20} \\
  DP3 & 9/20 & 28/80 & 2/20 \\
  Masquerade-3D & 0/20 & 47/80 & 2/20 \\
  \bottomrule
  \end{tabular}
\end{table}

\textbf{Real World.} We design three real-world tasks that require fine-grained manipulation skills as shown in Fig.~\ref{fig:tasks} (full protocols in Appendix~\ref{appendix:tasks}):
\begin{itemize}[leftmargin=10pt, labelindent=0pt]
    \item \textbf{Desk Cleanup:} the robot closes a laptop and a book without colliding, then picks up the mouse and places it aligned with the laptop. An episode succeeds only if all four steps succeed.
    \item \textbf{Candy Sorting:} the robot sorts eight randomly placed candy bags into two boxes by color. The bags are flat, cluttered, and in widely varying poses, which makes them hard to grasp. We score each bag rather than each episode: a bag fails if it lands in the wrong box or outside the boxes, or after more than three failed grasps, in which case we remove it and the policy continues.
    \item \textbf{K-Cup Coffee Making:} the most fine-grained and long-horizon task. The robot picks up the mug and places it under the machine, opens the lid, grasps a pod, inserts it, closes the lid, and presses the button, and succeeds only if all seven subtasks succeed.
\end{itemize}
As shown in Table~\ref{tab:real_result}, our method outperforms both baselines in all three tasks, by 51.7 points on average over DP3, the baseline with the higher average (43.8 points against the best baseline on each task). On K-Cup coffee making, DUO reaches 70\% while both baselines reach 10\%. Compared to DP3, co-training in Masquerade-3D improves Candy Sort by 23.8 points but degrades Desk Cleanup by 45.0 points. 

\begin{figure}[t]
  \centering
  \includegraphics[width=0.9\linewidth]{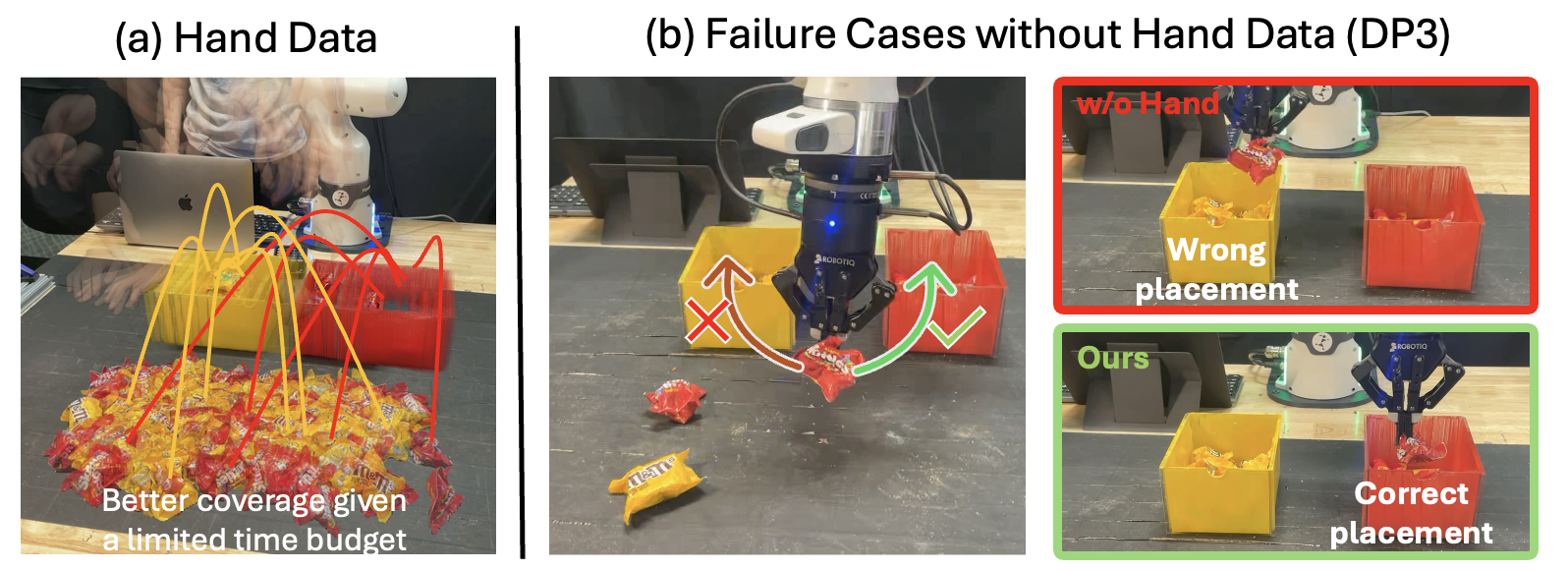}
  \caption{\textbf{Real-World Case Study on Candy Sorting.} (a) Hand data cover the cluttered bag configurations within a limited time budget. (b) Without hand data, DP3 places a bag in the wrong box (red); DUO places it in the box of matching color (green).}
  \label{fig:case_study_candy}
\end{figure}

\textbf{Baseline Failure Mode Analysis.} Fig.~\ref{fig:case_study} shows why each baseline fails. The hand data carry pose-estimation error but cover the workspace broadly within a limited time budget (Fig.~\ref{fig:case_study}(a)). Masquerade-3D co-trains a single policy on these noisy actions and gains coverage at the cost of precision: it misses the lid handle and collides during pod insertion (Fig.~\ref{fig:case_study}(b)). DP3, trained on robot data alone, is precise but lacks coverage: it misses the pod and the mug at poses unseen from the robot data (Fig.~\ref{fig:case_study}(c)). On Candy Sorting, it places bags in the wrong box (Fig.~\ref{fig:case_study_candy}) due to limited data coverage. DUO avoids both failures, taking coverage from the hand data and precision from the robot data.

\begin{figure}[t]
    \centering
    \includegraphics[width=0.7\linewidth]{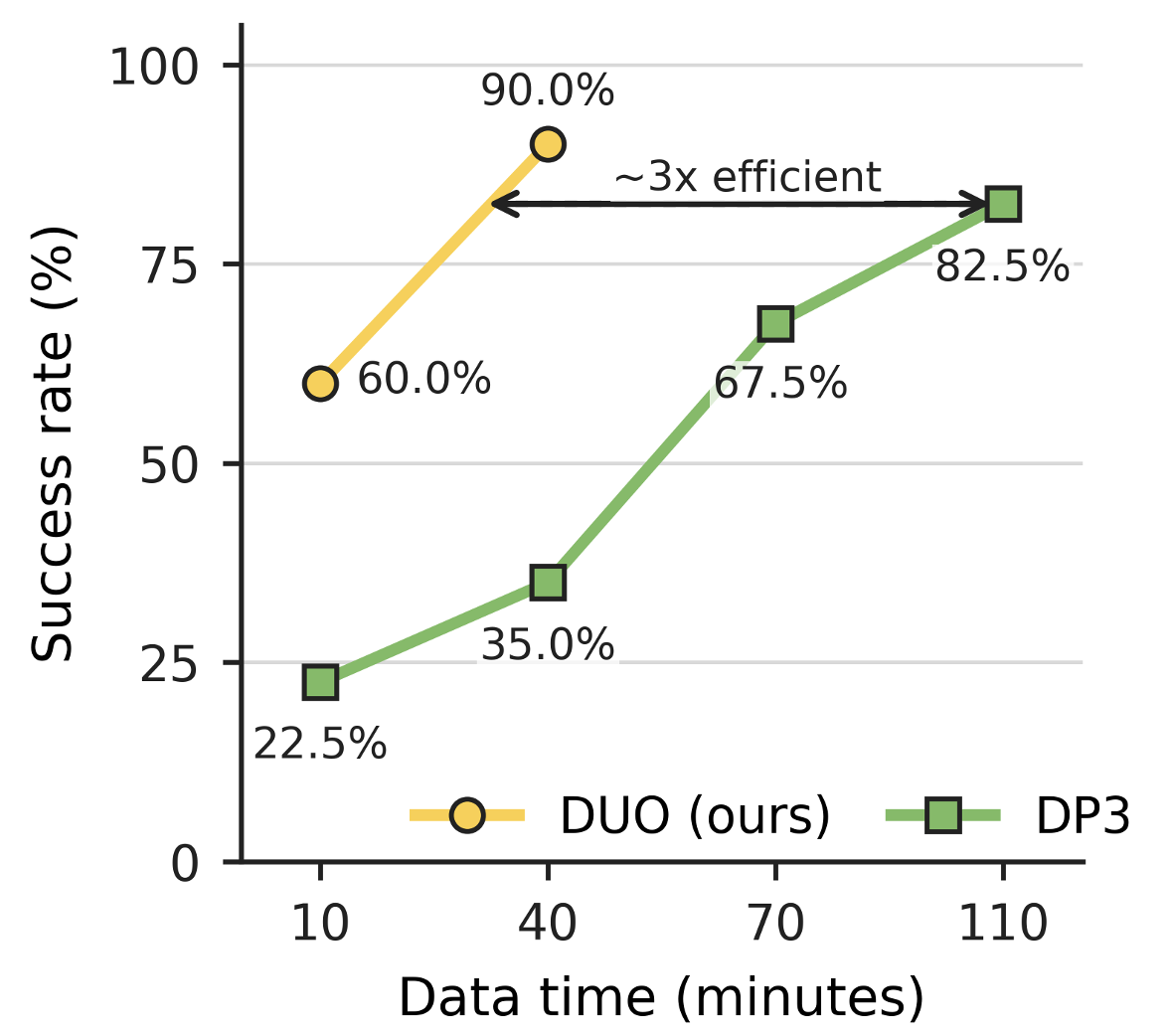}
    \caption{\textbf{Data Collection Time.} Success rate vs.\ wall-clock data-collection time on candy sorting.
  }
  \label{fig:time}
\end{figure}
\textbf{Data-Collection Efficiency.} We measure data-collection efficiency on candy sorting, whose wide range of candy-bag configurations demands a substantial amount of data to learn. For our method, we first spend 10 minutes collecting human hand demonstrations, yielding 216 human hand trajectories. We then train the hand policy and use it for human-in-the-loop data collection. After 30 additional minutes of HG-DAgger~\citep{kelly2019hgdagger}, our method achieves a 90\% success rate. Under the same 40-minute data collection budget, DP3 with robot demonstrations collected via SpaceMouse only reaches a 35\% success rate. DP3 needs another 70 minutes of robot data collection to reach 82.5\%, so DUO reaches a comparable success rate with 2.75$\times$ less collection time.

\subsection{Where Does Each Data Source Help?}
\label{sec:where}

\begin{figure*}[t]
  \centering
  \subfloat[Task]{\includegraphics[width=0.22\linewidth]{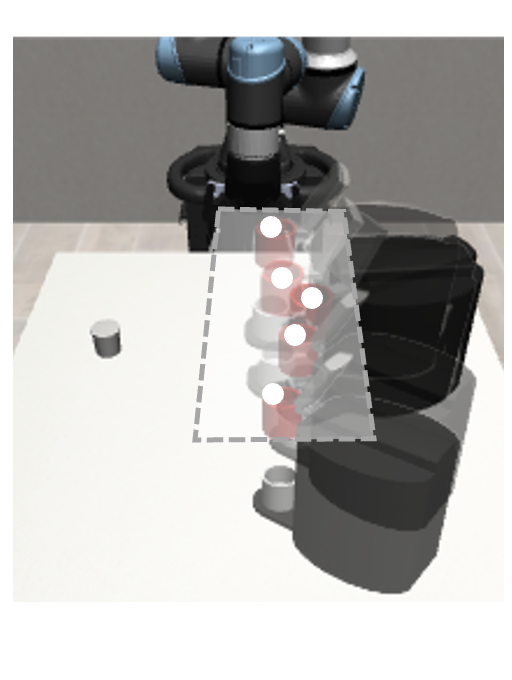}\label{fig:spatial_task}}\hfill
  \subfloat[Insertion Success Map]{\includegraphics[width=0.77\linewidth]{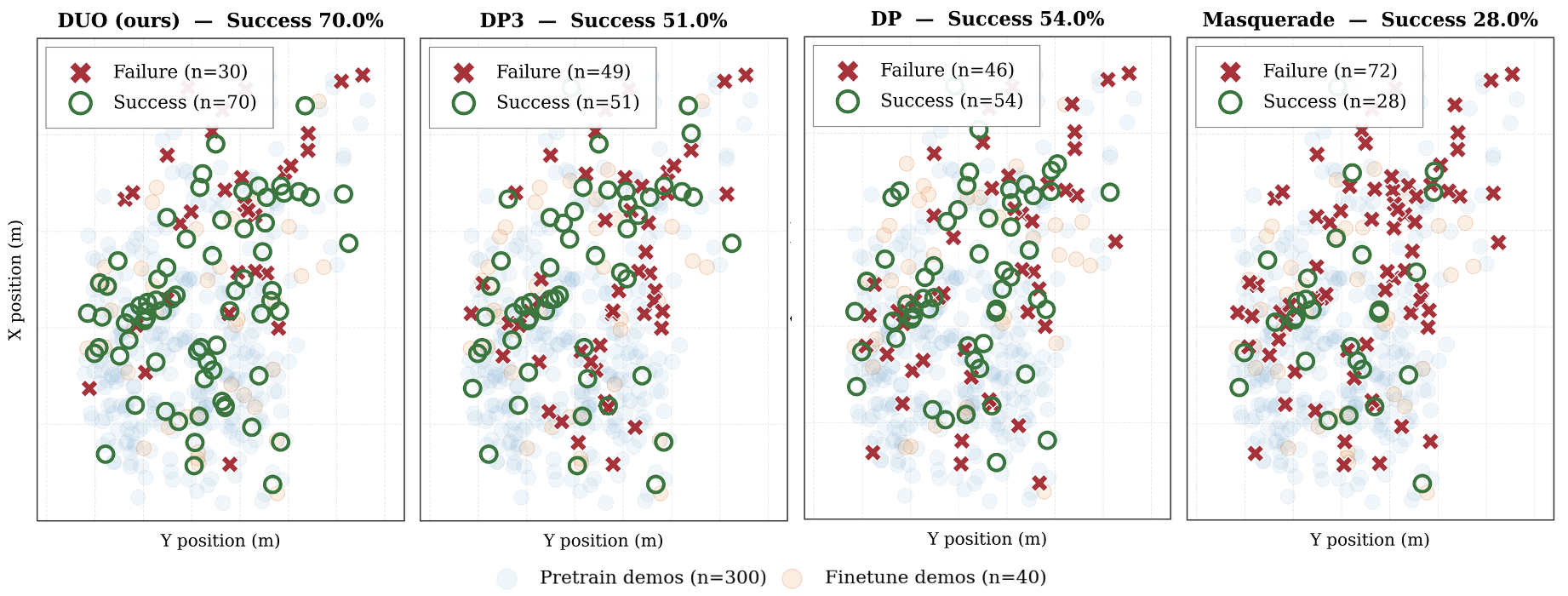}\label{fig:spatial_result}}
  \caption{\textbf{Spatial Generalization on Coffee D2.} \textbf{(a)} The coffee machine is placed at varying positions on the table. \textbf{(b)} Outcomes of 100 test episodes for each method, plotted at the coffee machine's position (one finetuning seed of Table~\ref{tab:sim_result}, 40 robot episodes), with pretraining and finetuning demonstrations in the background.}
  \label{fig:spatial}
\end{figure*}

Our hypothesis predicts that noisy cross-embodiment data contribute coverage where precision requirements are low, while robot data must govern precision-critical states.

\textbf{DUO turns noisy coverage into spatial generalization.} We test whether DUO completes the task, including the fine-grained insertion, at placements covered only by the noisy pretraining data. Figure~\ref{fig:spatial} plots each test episode on Coffee D2 at the position of the coffee machine. The 300 noisy pretraining demonstrations cover the workspace densely, while the 40 robot demonstrations are sparse. DUO succeeds on 70\% of 100 test placements, compared to 51\% for DP3, 54\% for Diffusion Policy, and 28\% for Masquerade. The failure patterns differ: DUO succeeds throughout the densely covered region, and its failures concentrate at the edge of the workspace, whereas the baselines also fail inside the covered region. This is the behavior our hypothesis predicts. The embodiment error in the pretraining data is tolerated in free space, so the hand policy brings the gripper to the holder from most of the workspace. Near the holder, the in-hand view is the same regardless of where the machine sits, so 40 robot demonstrations suffice to learn the local insertion. The baselines instead rely on a single finetuned policy, which must learn the precise insertion from only 40 robot episodes while also covering the approach.

\textbf{DUO is robust to where the embodiment error lies.} Wherever we inject pose noise into the pretraining data, DUO stays within 1.0 points of its clean success rate, while a co-trained DP3 loses up to 9.6 points (Appendix~\ref{appendix:extra}).

\subsection{Does the Gate Find Precision-Critical States?}
\label{sec:gate}

\begin{figure*}[t]
  \centering
  \includegraphics[width=\textwidth]{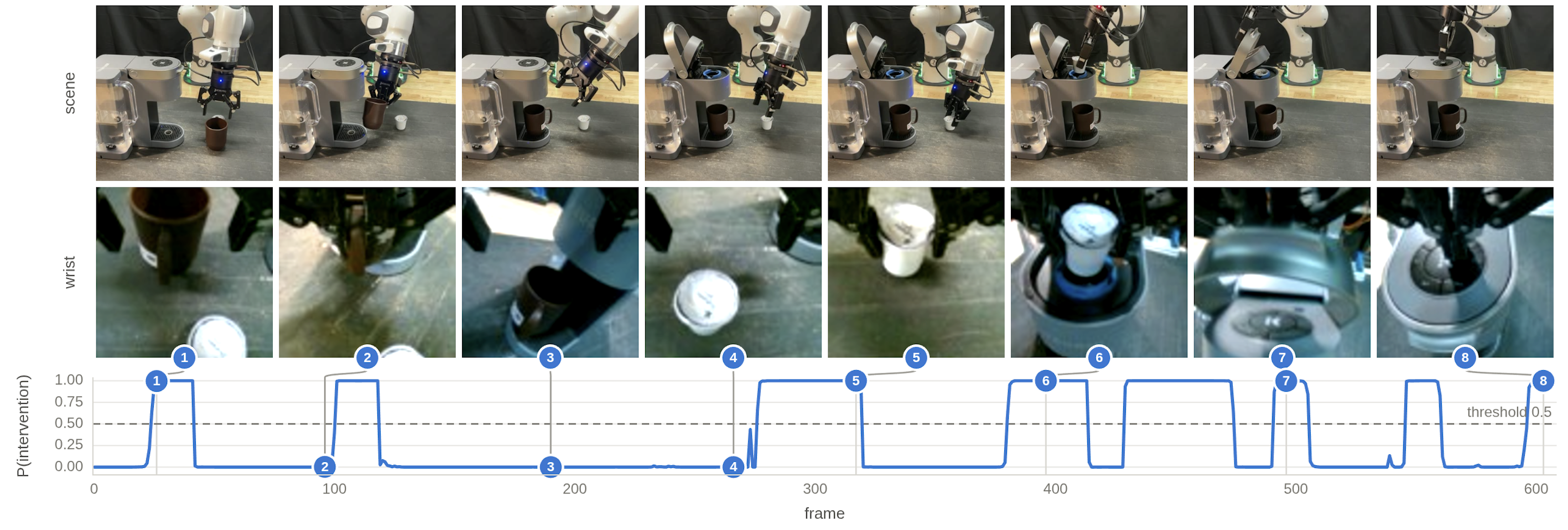}
  \caption{\textbf{Learned Gate on a Real-World K-Cup Coffee Trajectory.} Top: external-camera views. Middle: in-hand views. Bottom: the gate's predicted intervention probability over time. Control passes from $\pi_{\text{hand}}$ to $\pi_{\text{robot}}$ when the probability exceeds 0.5. The gate switches on at the contact-rich steps, such as grasping the mug and the pod, inserting the pod, and pressing the button, and stays near zero during free-space motion.}
  \label{fig:gate_switch}
\end{figure*}

\begin{table}[t]
  \centering
  \caption{Gate Analysis}
  \label{tab:gate}
  \footnotesize
  \setlength\tabcolsep{5pt}
  \begin{tabular}{l c c c}
  \toprule
  \textbf{Gate} & \textbf{Seen} & \textbf{Unseen} & \textbf{F1} \\
  \midrule
  Always hand & 0 & 0 & -- \\
  Always robot & 41{\scriptsize$\pm$1} & 33{\scriptsize$\pm$3} & -- \\
  Distance$^\dagger$ & 49{\scriptsize$\pm$1} & 44{\scriptsize$\pm$1} & 0.70 \\
  Privileged phase$^\dagger$ & 72{\scriptsize$\pm$7} & 73{\scriptsize$\pm$6} & 1.00 \\
  \midrule
  \textbf{Learned (ours)} & \textbf{78}{\scriptsize$\pm$6} & \textbf{79}{\scriptsize$\pm$2} & 0.85 \\
  \bottomrule
  \end{tabular}
\par\vspace{2pt}\parbox{\linewidth}{\footnotesize Success rate (\%, mean $\pm$ std over 3 seeds) on placements inside (Seen) and outside (Unseen) the region covered by 18 robot episodes collected by a human operator intervening on $\pi_{\text{hand}}$. F1 is against the privileged phase label.}
\end{table}

To test whether the states where the gate hands over control are the precision-critical ones, we keep both trained branches fixed and replace only the gate (Table~\ref{tab:gate}). Always-hand and always-robot use a single branch throughout. The privileged phase gate switches on the task-phase label from MimicGen's subtask segmentation~\citep{mandlekar2023mimicgen}, which uses privileged simulator state, and the distance gate hands control to the robot branch when the gripper is within 10\,cm of the pod or 20\,cm of the pod holder, using privileged object poses. We evaluate on the spatial-generalization layout (Fig.~\ref{fig:spatial}), separating placements inside (seen) and outside (unseen) the region covered by the robot data. The learned gate matches the privileged phase gate (78 vs.\ 72\% seen, 79 vs.\ 73\% unseen) and agrees with it on most timesteps (F1 0.85). A phase label marks where the task changes, not where $\pi_{\text{hand}}$ actually fails. The learned gate is trained on interventions, which mark the latter, so it can hand over control earlier or later as the hand policy's errors require. Its success does not drop on unseen placements, because the in-hand gate only needs to recognize local geometry. In contrast, the distance gate switches to the robot branch too early, so it behaves much like always-robot and inherits its weakness: the robot branch must also cover approach motion, and success falls from 49\% to 44\% on unseen placements (41\% to 33\% for always-robot). Always-hand never completes the task, confirming that the human prior alone cannot handle the precision-critical states.

\begin{table}[t]
  \centering
  \caption{Learned Gate on Held-Out Real-World Trajectories}
  \label{tab:gate_real}
  \footnotesize
  \setlength\tabcolsep{3pt}
  \begin{tabular}{l c c c c c c}
  \toprule
  \textbf{Task} & \textbf{Interv.} & \textbf{Acc.} & \textbf{Prec.} & \textbf{Rec.} & \textbf{F1} & \textbf{AUROC} \\
  \midrule
  K-Cup Coffee & 51.7 & 94.1 & 97.4 & 91.0 & 94.1 & 0.991 \\
  Desk Cleanup & 29.4 & 96.3 & 97.3 & 90.0 & 93.5 & 0.992 \\
  Candy Sort & 41.1 & 97.6 & 98.5 & 95.6 & 97.0 & 0.998 \\
  \bottomrule
  \end{tabular}
  \par\vspace{2pt}\parbox{\linewidth}{\footnotesize Gate predictions against intervention labels (\%, except AUROC). Interv.: percentage of intervention frames ($i_t=1$) among all.}
\end{table}

\textbf{Real-world gate.} On held-out real-world trajectories, the learned gate predicts the intervention labels with 94--98\% accuracy, at least 97\% precision, and an AUROC of at least 0.99 on all three tasks (Table~\ref{tab:gate_real}). Fig.~\ref{fig:gate_switch} shows the gate along one K-Cup trajectory, where misclassifications occur only near switching boundaries and shift the handover by a short offset.

\subsection{Ablations}
\label{sec:ablation}

\begin{table}[t]
  \centering
  \caption{Ablation Study}
  \label{tab:ablation}
  \footnotesize
  \setlength\tabcolsep{3.5pt}
  \begin{tabular}{l l c c}
    \toprule
    \textbf{Component} & \textbf{Variant} & \textbf{Success} & $\boldsymbol{\Delta}$ \\
    \midrule
    \textbf{Dual actors} & Single policy, SFT & 73.0{\scriptsize$\pm$4.5} & $-12.7$ \\
    (Sec.~\ref{sec:dual}) & Single policy, co-training & 71.0{\scriptsize$\pm$1.4} & $-14.7$ \\
    \addlinespace[2pt]
    \textbf{Human prior} & No human pretraining & 18.3{\scriptsize$\pm$2.5} & $-67.3$ \\
    (Sec.~\ref{sec:canon}) & No canonicalization $\phi$ & 54.0{\scriptsize$\pm$2.4} & $-31.7$ \\
    & Point cloud $\rightarrow$ image & 39.3{\scriptsize$\pm$4.9} & $-46.3$ \\
    & Multi-view $\rightarrow$ single-view & 62.7{\scriptsize$\pm$3.7} & $-23.0$ \\
    \midrule
    & \textbf{Full DUO} & \textbf{85.7}{\scriptsize$\pm$1.7} & -- \\
    \bottomrule
    \end{tabular}
\par\vspace{2pt}\parbox{\linewidth}{\footnotesize Success rates (\%), mean $\pm$ std over three seeds, under the protocol of Table~\ref{tab:sim_result}: final checkpoint (epoch 600), 100 unseen configurations. All rows finetune on the same 40 intervention episodes collected by a human operator, so absolute values differ from the scripted-intervener runs of Table~\ref{tab:sim_result}. $\Delta$ is the difference to Full DUO in percentage points, computed from unrounded means.}
\end{table}

\begin{table}[t]
  \centering
  \caption{Locality and Equivariance of the Robot Actor}
  \label{tab:globallocal}
  \footnotesize
  \setlength\tabcolsep{2.5pt}
  \begin{tabular}{l l c c c}
    \toprule
    & & \multicolumn{2}{c}{\textbf{Coffee D2}} & \textbf{Stack Three} \\
    \cmidrule(lr){3-4}\cmidrule(lr){5-5}
    \textbf{Obs.} & \textbf{Actions} & \textbf{Seen} & \textbf{Unseen} & \textbf{In-dist} \\
    \midrule
    PCD & Absolute & \phantom{0}6.0{\scriptsize$\pm$0.8} & \phantom{0}3.3{\scriptsize$\pm$0.5} & \phantom{0}2.3{\scriptsize$\pm$0.5} \\
    PCD & Relative & 10.3{\scriptsize$\pm$3.7} & \phantom{0}4.0{\scriptsize$\pm$2.8} & 13.3{\scriptsize$\pm$2.6} \\
    Inhand & Absolute & 50.0{\scriptsize$\pm$7.1} & 29.7{\scriptsize$\pm$3.7} & 24.7{\scriptsize$\pm$4.5} \\
    \midrule
    \textbf{Inhand} & \textbf{Relative} & \textbf{74.7}{\scriptsize$\pm$2.4} & \textbf{77.7}{\scriptsize$\pm$3.3} & \textbf{66.3}{\scriptsize$\pm$3.1} \\
    \bottomrule
  \end{tabular}
  \par\vspace{2pt}\parbox{\linewidth}{\footnotesize Success rates (\%) of DUO with the robot actor's observation and action frame varied. Mean $\pm$ std over 3 seeds. The last row is full DUO. PCD: world-frame scene point cloud without in-hand cameras. Inhand: in-hand cameras. Coffee D2: 18 robot demonstrations in a sub-region of coffee-machine poses, 200 evaluation episodes split by whether the placement lies inside (Seen) or outside (Unseen) that region. Stack Three: the 40 robot episodes and 100 evaluation episodes of Table~\ref{tab:sim_result}; the last row is the Table~\ref{tab:sim_result} run. The Coffee D2 runs are trained and evaluated separately from Tables~\ref{tab:gate} and~\ref{tab:ablation}.}
\end{table}

As shown in Table~\ref{tab:ablation}, we ablate each component. \textbf{Dual actors.} Replacing the dual actors with a single policy trained by SFT or co-training drops performance by 12.7 and 14.7 percentage points, consistent with Section~\ref{sec:where}: co-training lets the embodiment error leak into precision-critical states, and SFT risks forgetting the human prior. \textbf{Human prior.} Without human pretraining, success drops by 67.3 points, since 40 robot demonstrations alone cannot cover the workspace. Without the canonicalization $\phi$, which renders the target gripper into the pretraining observations, it drops by 31.7 points because of the visual gap between pretraining and deployment observations. Replacing the 3D point cloud with image input drops it by 46.3 points since the hand policy has no in-hand view and relies on 3D geometry to act from human videos. With a single-view point cloud, DUO still reaches 62.7\%, far above training without human data (18.3\%). 

\textbf{Local robot actor and equivariance.} Table~\ref{tab:globallocal} tests whether the robot actor must be local. Proposition~\ref{prop:equiv} requires $\pi_{\text{robot}}$ to observe the local in-hand view and act in the relative frame, and we break each condition by giving it the global observation (global point cloud) or absolute actions instead. To isolate the effect of equivariance, on Coffee D2 the robot actor trains on 18 demonstrations from a sub-region of coffee-machine poses and is evaluated on placements inside (seen) and outside (unseen) this region. On seen placements, the global observation lowers success from 74.7\% to 10.3\%, and absolute actions lower it to 50.0\%. On unseen placements, which the robot data do not cover, DUO keeps its success (77.7\%), whereas the absolute-action variant falls to 29.7\% and both global-observation variants to 4.0\% or less. Stack Three D1 shows the same ranking. Without either condition, the robot actor's input or output depends on the placement, so the sparse robot data would have to cover placement variation itself.

\section{Conclusion and Limitations}

Cross-embodiment data is most useful where precision requirements are low, while target-embodiment data is most valuable near precision-critical states. In this paper, we propose DUO Policy, a dual-actor policy that puts this principle into practice. DUO Policy switches, via a gate learned from human interventions, between a hand policy pretrained on human data for free-space motion and a robot policy trained on sparse in-hand robot data for precise corrections. DUO outperforms all baselines in 8 of 12 simulated settings and in all three real-world tasks, exceeding the baseline with the highest average success by 12.9 percentage points in simulation and 51.7 points in the real world, and on candy sorting it matches robot-only training with $2.75\times$ less data-collection time. Its learned gate matches a privileged phase gate, and it generalizes beyond the region covered by the robot data.

\textbf{Limitations.} The hand policy consumes point clouds, so collecting human videos and deploying the policy require at least one calibrated RGB-D camera. Future work will reduce the reliance on depth, for example with monocular depth estimation, and scale DUO Policy to open-source egocentric datasets~\citep{grauman2022ego4d, hoque2025egodex}. In addition, the robot data are collected by intervening on a specific hand policy, so retraining or replacing that policy may require new robot data. We plan to explore training the gate and robot actor on interventions from multiple hand policies and multi-task datasets, which would allow the robot data to be reused across hand-policy updates.

\appendices

\section{Proof of Proposition~\ref{prop:equiv}}
\label{appendix:proof}
\textbf{Setup.} Let $T_t \in SE(3)$ be the world-frame gripper pose. We embed $T_g \in SE(2)$ in $SE(3)$ as a rotation about the gravity axis followed by a horizontal translation; it acts on the state $s_t$ by left-multiplying every scene pose and the gripper pose, so $T_t \mapsto T_g T_t$. We assume that the in-hand view $o_t^{\text{local}} = h(s_t)$ depends on $s_t$ only through the gripper-frame state $T_t^{-1} \cdot s_t$.

\textbf{Proof.} In the transformed state $T_g \cdot s_t$ the gripper pose is $T_g T_t$, so the gripper-frame state is $(T_g T_t)^{-1}\, T_g \cdot s_t = T_t^{-1} \cdot s_t$, unchanged. Hence $h(T_g \cdot s_t) = h(s_t)$. The gate sees $s_t$ only through $h$, which gives the first identity in Eq.~(\ref{eq:equiv}). If $\gamma=1$, $\pi_{\text{robot}}$ receives the same input and predicts the same relative motions $\Delta_t^{(k)}$, $k=1,\dots,K$, which DUO executes from the current pose as $a_{t+k} = T_t\,\Delta_t^{(k)}$. In the transformed state these become $(T_g T_t)\,\Delta_t^{(k)} = T_g\,a_{t+k}$, so the whole chunk transforms by $T_g$, which is the second identity. For a stochastic $\pi_{\text{robot}}$, the equality holds in distribution. \hfill $\square$

\textbf{Remarks.} (1) The gripper-frame assumption holds approximately in practice, since the in-hand view may include background that does not move with $T_g$. (2) If $\pi_{\text{robot}}$ predicts absolute actions, its target must move from $T^\star_t$ to $T_g T^\star_t$ when the scene moves, but its in-hand view does not change. It therefore needs the end-effector pose as an extra input and must learn this mapping from the sparse robot data. (3) If $\pi_{\text{robot}}$ observes the global point cloud, its input changes with the placement, so it must learn from data how its output should change. In both cases the equivariance is no longer built in. The gate always uses the in-hand view and stays invariant (Table~\ref{tab:globallocal}). (4) When $\gamma=0$, the $SO(2)$ augmentation of $\pi_{\text{hand}}$ (Sec.~\ref{sec:dual}) makes it approximately equivariant to rotations about the gravity axis. This equivariance is learned, not built into the architecture, and excludes translations.

\FloatBarrier
\section{Implementation and Task Details}
\label{appendix:impl}

\subsection{Human Data Processing}
\label{appendix:human}
\begin{figure}[t]
    \centering
    \includegraphics[width=\linewidth]{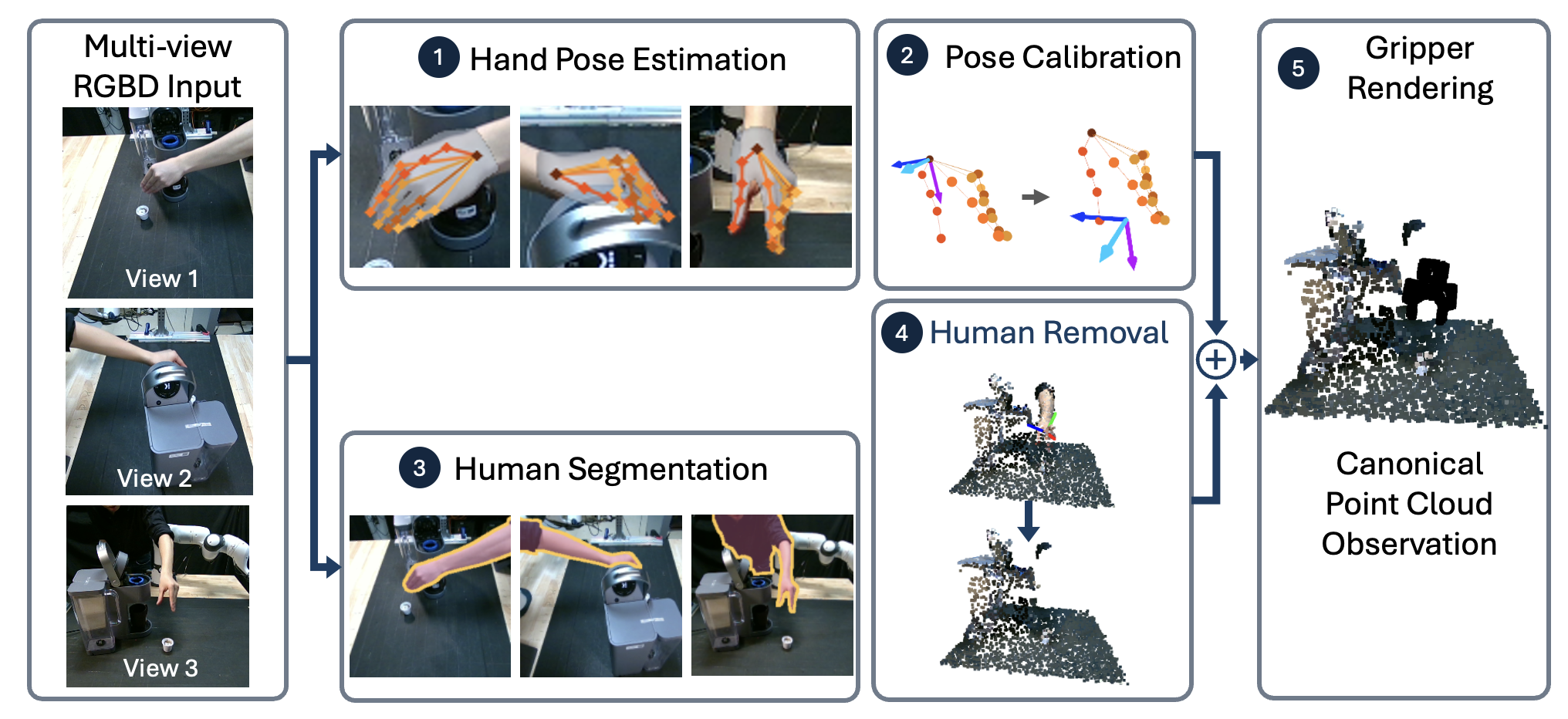}
    \caption{\textbf{Human Data Preprocessing Pipeline (the canonicalization map $\phi$).} Numbers mark the steps described in Appendix~\ref{appendix:human}.}
    \label{fig:pipeline}
\end{figure}

\begin{figure}[t] \centering \includegraphics[width=0.7\linewidth]{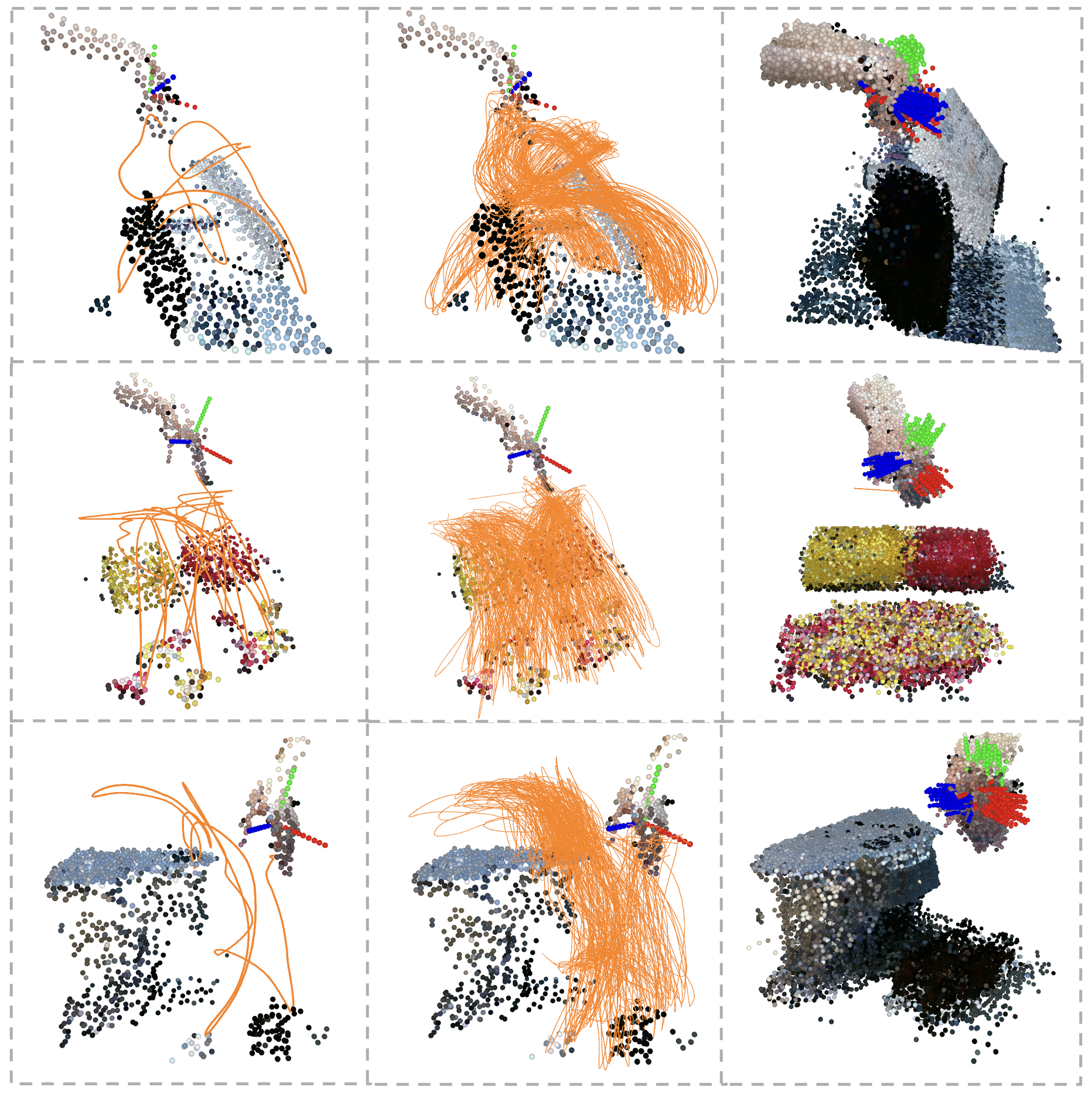} \caption{\textbf{Coverage of the Human Data.} Processed point-cloud observations for the three real-world tasks (rows, top to bottom: K-Cup Coffee Making, Candy Sorting, Desk Cleanup). \textbf{Left:} one human-hand trajectory (orange), retargeted to the gripper. \textbf{Middle:} all retargeted trajectories, which span the workspace. \textbf{Right:} initial scenes and gripper poses of all demonstrations overlaid, showing the range of object layouts and start poses that the human data covers.} \label{fig:hand_pcd} \end{figure}

\textbf{Preprocessing Pipeline.} Figure~\ref{fig:pipeline} illustrates the canonicalization map $\phi$ described in Section~\ref{sec:canon}, and Fig.~\ref{fig:hand_pcd} visualizes how much of the workspace and how many scene layouts the human data covers in the real-world tasks. From multi-view RGB-D input, (1) WiLoR estimates the hand pose in each view, and (2) the poses are converted to a fingertip-based gripper pose that serves as the cross-embodiment action. In parallel, (3) SAM3 segments the human body, and (4) the remaining pixels are back-projected into a human-free point cloud. (5) The target gripper is rendered into this point cloud at the fingertip-derived pose, so human and robot observations match.

\textbf{Feasibility-aware Human Data Collection.} Some hand motions that are easy for a human cannot be executed by the robot because of joint limits or inverse-kinematics failures. Our collection interface therefore runs WiLoR online, retargets the estimated hand pose to a target gripper pose, and solves inverse kinematics for the robot in real time. Whenever a motion would lead to an infeasible configuration, the interface immediately alerts the demonstrator, who adjusts the motion during the demonstration, so few infeasible trajectories are recorded. All real-world experiments use a Franka arm with a Robotiq gripper, RealSense D455 side-view and D405 in-hand cameras, and a SpaceMouse for interventions and DP3 demonstrations.

\subsection{Real-World Tasks}
\label{appendix:tasks}

\textbf{Desk Cleanup}: The robot must (1) fully close the laptop, (2) fully close the book, (3) pick up the mouse, and (4) place it next to the laptop with the correct alignment, without colliding with either object. An episode succeeds only if all four steps succeed. Data: 25\,min of human and 50\,min of robot data.

\textbf{Candy Sorting}: Eight candy bags and two boxes are placed randomly, and the robot must place each bag in the box of its color. A bag counts as a failure if it is placed in the wrong box, lands outside the boxes, or the robot fails to grasp it more than three times; in the last case the bag is removed and the policy continues with the remaining bags. We report the fraction of bags sorted correctly. Data: 10\,min of human and 30\,min of robot data.

\textbf{K-Cup Coffee Making}: The robot must (1) pick up the mug, (2) place it under the machine, (3) open the lid, (4) grasp a coffee pod, (5) insert the pod, (6) close the lid, and (7) press the brew button. An episode succeeds only if all seven subtasks succeed. Data: 38\,min of human and 120\,min of robot data.

\subsection{Hyperparameters}
\label{appendix:hyper}

Table~\ref{tab:hyper} lists the settings of DUO, shared by the simulation and real-world experiments unless noted. Both actors share the Diffusion Policy~\citep{diffpo} architecture; $\pi_{\text{robot}}$ is a separate network trained from scratch (apart from its pretrained ResNet-18 image encoder) and shares no parameters with $\pi_{\text{hand}}$.

\begin{table}[!h]
  \centering
  \caption{Hyperparameters and Settings}
  \label{tab:hyper}
  \footnotesize
  \setlength\tabcolsep{4pt}
  \begin{tabular}{@{}>{\raggedright\arraybackslash}p{0.27\linewidth}>{\raggedright\arraybackslash}p{0.69\linewidth}@{}}
  \toprule
  \multicolumn{2}{@{}l}{\textit{Both actors}} \\
  Denoiser & 1D conv.\ UNet (256/512/1024), FiLM; 70.5M \\
  Diffusion & DDPM, $\epsilon$-pred., 100 steps \\
  Horizons (obs./pred./exec.) & 1 hand, 2 robot / 16 / 8 \\
  Training & batch 256, lr $10^{-4}$ cosine \\
  \midrule
  \multicolumn{2}{@{}l}{\textit{Hand actor $\pi_{\text{hand}}$ (79.4M, frozen in finetuning)}} \\
  Input & 1024-point cloud (xyz+RGB) \\
  Output & absolute world-frame pose \\
  Augmentation & $SO(2)$ yaw, $\pm180^\circ$ sim / $\pm45^\circ$ real \\
  \midrule
  \multicolumn{2}{@{}l}{\textit{Robot actor $\pi_{\text{robot}}$ and gate $\gamma$ (81.8M)}} \\
  Input & in-hand RGB with Pretrained ResNet-18 \\
  Output & gripper-frame relative poses \\
  Gate & BCE on $i_t$, $\lambda=0.1$, threshold 0.5; queried every 8-step chunk (sim) / every step (real) \\
  Sampling & interventions : others = 1 : 1 \\
  \bottomrule
  \end{tabular}
\end{table}

\section{Additional Results}
\label{appendix:extra}

\begin{figure}[t]
    \centering
    \includegraphics[width=\linewidth]{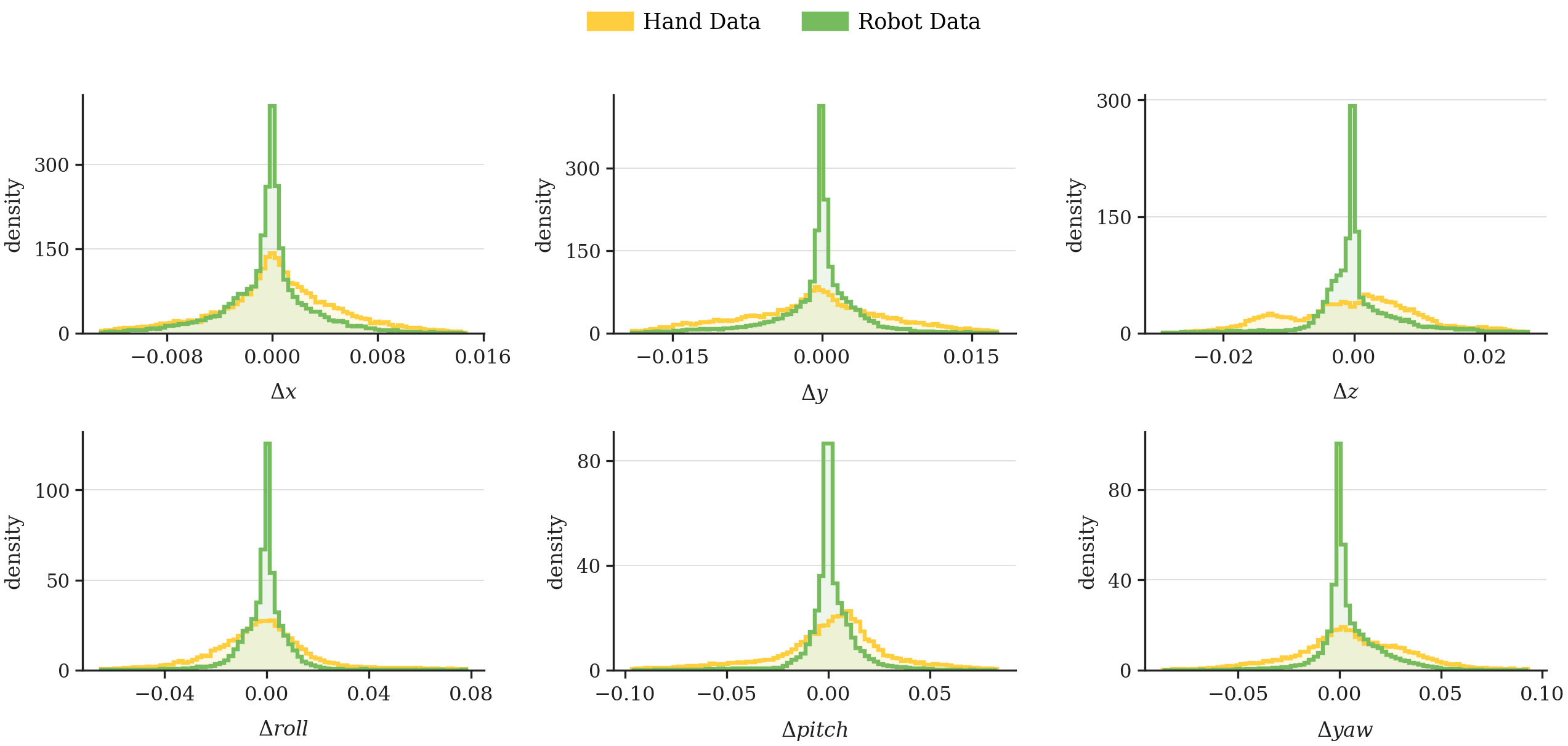}
    \caption{\textbf{Action Distribution Comparison.} Human hand actions vs.\ robot actions collected with a SpaceMouse (without interventions).}
    \label{fig:action_dist}
\end{figure}

\textbf{Decoupled Action Normalization.} Human and robot actions differ in scale (Fig.~\ref{fig:action_dist}). Robot actions collected by teleoperation are sharply peaked around zero, since robot trajectories consist of many small adjustments, whereas a human moving their own hand takes larger steps. A joint normalization $\nu(a) = (a - \mu_{\text{joint}}) / \sigma_{\text{joint}}$ would let the dominant human variance ($\sigma_{\text{hand}} \gg \sigma_{\text{robot}}$) compress the robot actions into a narrow band near zero and weaken the learning signal for fine corrections. Each actor in DUO therefore uses its own statistics, $(\mu_{\text{hand}}, \sigma_{\text{hand}})$ and $(\mu_{\text{robot}}, \sigma_{\text{robot}})$.

\begin{wrapfigure}{r}{0.45\linewidth}
  \centering
  \vspace{-10pt}
  \includegraphics[width=\linewidth]{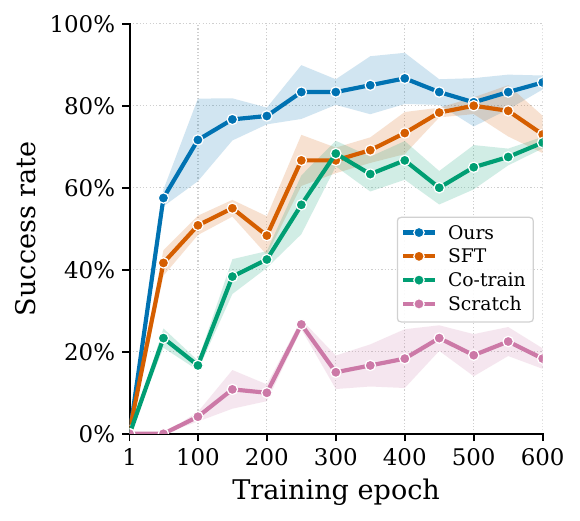}
  \caption{\textbf{Success during Finetuning on Coffee D2.} Evaluated every 50 epochs.}
  \label{fig:robust_conv}
  \vspace{-10pt}
\end{wrapfigure}
\textbf{Finetuning Efficiency.} Fig.~\ref{fig:robust_conv} evaluates the variants of Table~\ref{tab:ablation} every 50 finetuning epochs on Coffee D2; the values at epoch 600 are those of Table~\ref{tab:ablation}. DUO exceeds 70\% after 100 epochs, a level SFT reaches after about 400 epochs and co-training only at epoch 600. Because $\pi_{\text{hand}}$ is frozen and already handles the free-space motion, the robot actor only has to learn the local corrections. A single policy must instead adapt its entire behavior to the robot data: SFT improves more slowly and declines after epoch 500, and co-training plateaus between 60\% and 70\%, as its noisy hand actions keep competing with the robot corrections. Training from scratch stays below 30\% throughout. DUO matches or exceeds every variant at every evaluated epoch, so its advantage does not depend on the choice of checkpoint.

\textbf{Precision Sensitivity by State.} We design this test to measure how much action error each state tolerates, i.e., the tolerance $\varepsilon(s)$ of the task itself rather than of a learned policy. At 15 states from each of 20 demonstrations, split by the privileged phase label, we perturb 8 expert actions in 8 directions, let the expert continue, and record the drop in success. On Coffee D2, shifting the expert's actions by 2\,cm lowers its success rate by 9.4 percentage points when the shift occurs in a precision-critical state, but by only 0.2 points when it occurs in free space. Stack Three D1 shows the same pattern, with drops of 2.3 and 0.1 points. 

\textbf{Noise Localization.} To test whether DUO depends on where the embodiment error lies, we inject the $\pm$2\,cm pose noise of Table~\ref{tab:sim_result} into the pretraining data only in free space, only near precision-critical states, or everywhere, and finetune every variant on one fixed set of 40 robot episodes so that only the pretraining data change (Table~\ref{tab:noise_loc}). DUO stays within 1.0 points of its clean success rate of 73.7\% in every condition; its 4.3-point gain under critical-only noise is within seed variance. DP3-cotrain, a single DP3 policy co-trained on the pretraining and robot data, loses 2.0 points under free-space noise, 6.6 under critical-only noise, and 9.6 under noise everywhere, and DUO exceeds it by 29 or more points in every row. The precision-sensitivity test above isolates where the error matters more directly; this experiment shows that DUO's success does not depend on it. The gate hands precision-critical states to $\pi_{\text{robot}}$, which never trains on the noisy pretraining data, so the error in $\pi_{\text{hand}}$ acts mainly in free space, where it is tolerated.

\begin{table}[t]
  \centering
  \caption{Noise Localization}
  \label{tab:noise_loc}
  \footnotesize
  \setlength\tabcolsep{5pt}
  \begin{tabular}{l c c}
  \toprule
  \textbf{Noise location} & \textbf{DUO} & \textbf{DP3-cotrain} \\
  \midrule
  None (clean) & 73.7{\scriptsize$\pm$2.5} & 44.3{\scriptsize$\pm$4.7} \\
  Free space only & 72.7{\scriptsize$\pm$3.7} & 42.3{\scriptsize$\pm$2.1} \\
  Critical states only & 78.0{\scriptsize$\pm$1.6} & 37.7{\scriptsize$\pm$5.6} \\
  Everywhere & 73.3{\scriptsize$\pm$5.8} & 34.7{\scriptsize$\pm$4.1} \\
  \bottomrule
  \end{tabular}
  \par\vspace{2pt}\parbox{\linewidth}{\footnotesize Success rates (\%), mean $\pm$ std over 3 seeds with 100 evaluation episodes each, with the $\pm$2\,cm pose noise of Table~\ref{tab:sim_result} injected into the pretraining data. All rows finetune on one fixed set of 40 robot episodes, collected by the scripted intervener on the original DUO pretrain. Only the noise in the pretraining data changes.}
\end{table}

\ifanon\else
\section*{Acknowledgment}
This work was supported in part by the Office of Naval Research under Grant N000142412784.
\fi

\bibliographystyle{IEEEtran}
\bibliography{main}

\end{document}